\documentclass[10pt,twocolumn,letterpaper]{article}

\usepackage{iccv}
\usepackage{epsfig}
\usepackage{graphicx}
\usepackage{amsmath}
\usepackage{amsthm}
\usepackage{amssymb}
\usepackage{subfig}
\usepackage{multirow}
\usepackage{color}
\usepackage{tabu}
\usepackage{array,times}
\usepackage[ruled,vlined]{algorithm2e}
\usepackage[american]{babel}

\newcommand{\squishlist}{
	\begin{list}{$\bullet$}
		{ \setlength{\itemsep}{0pt}
			\setlength{\parsep}{2pt}
			\setlength{\topsep}{2pt}
			\setlength{\partopsep}{0pt}
			\setlength{\leftmargin}{1em}
			\setlength{\labelwidth}{1em}
			\setlength{\labelsep}{0.5em} } }
	\newcommand{\squishend}{
	\end{list}  }

\newcommand{\red}[1]{\textcolor[rgb]{1, 0, 0}{#1}}

\newcommand{\sz}{\fontsize{6pt}{\lineskip}\selectfont}

\newcommand{\mat}[1]{\mbox{\boldmath{$#1$}}}
\newcommand{\ldbracket}{{[\kern-0.17em[}}
\newcommand{\rdbracket}{{]\kern-0.17em]}}

\usepackage[pagebackref=false,breaklinks=true,letterpaper=true,colorlinks,citecolor=blue,bookmarks=false]{hyperref}

\iccvfinalcopy %

\begin{document}

\title{When Unsupervised Domain Adaptation Meets Tensor Representations\thanks{Appearing in Proc.\ Int. Conf. Computer
    Vision (ICCV2017).
    HL and LZ contributed equally.
    This work was done when HL, LZ, and KX were visiting The University of Adelaide. ZC is the correspondence author.
  }
}

\author{Hao~Lu$^\dagger$,~Lei~Zhang$^\ddagger$,~Zhiguo~Cao$^\dagger$,~Wei~Wei$^\ddagger$,~Ke~Xian$^\dagger$,
  Chunhua~Shen$^\mathsection$, ~Anton~van~den~Hengel$^\mathsection$\\
$^\dagger$Huazhong Unviersity of Science and Technology,  China\\
$^\ddagger$Northwestern Polytechnical University, China\\
$^\mathsection$The University of Adelaide,  Australia\\
  e-mail: \tt \{poppinace,zgcao\}@hust.edu.cn
%
}

\maketitle

\begin{abstract}
Domain adaption (DA) allows machine learning methods trained on data sampled from one distribution to be applied to data sampled from another.  It is thus of great practical importance to the application of such methods. Despite the fact that tensor representations are widely used in Computer Vision to capture multi-linear relationships that affect the data, most existing DA methods are applicable to vectors only. This renders them incapable of reflecting and preserving important structure in many problems. We thus propose here a learning-based method to adapt the source and target tensor representations directly, without vectorization. In particular, a set of alignment matrices is introduced to align the tensor representations from both domains into the invariant tensor subspace. These alignment matrices and the tensor subspace are modeled as a joint optimization problem and can be learned adaptively from the data using the proposed alternative minimization scheme. Extensive experiments show that our approach is capable of preserving the discriminative power of the source domain, of resisting the effects of label noise, and works effectively for small sample sizes, and even one-shot DA. We show that our method outperforms the state-of-the-art on the task of cross-domain visual recognition in both efficacy and efficiency, and particularly that it outperforms all comparators when applied to DA of the convolutional activations of deep convolutional networks.
\end{abstract}

\tableofcontents
\clearpage


%
\section{Introduction}

The difficulty of securing an appropriate and exhaustive set of training data, and the tendency for the domain of application to drift over time, often lead to variations between the distributions of the training (source) and test (target) data. In Machine Learning this problem is labeled \mbox{\textit{domain mismatch}}. Failing to model such a distribution shift may cause significant performance degradation. Domain adaptation (DA) techniques capable of addressing this problem of distribution shift have thus received significant attention recently~\cite{patel2015visual}.

The assumption underpinning DA is that, although the domains differ, there is sufficient commonality to support adaptation. Many approaches have modeled this commonality by learning an invariant subspace, or set of subspaces~\cite{aljundi2015landmarks,fernando13sa,gong12geodesic,gopalan2011domain}. These methods are applicable to vector data only, however. Applying these methods to structured high-dimensional representations (e.g., convolutional activations), thus requires that the data be vectorized first. Although this solves the algebraic issue, it does not solve the underlying problem.

Tensor arithmetic is a generalization of matrix and vector arithmetic, and is particularly well suited to representing multi-linear relationships that neither vector nor matrix algebra can capture naturally~\cite{vasilescu2003multilinear}. The higher-order statistics of a vector-valued random variables are most naturally expressed as tensors, for instance. The power of tensor representations has also been demonstrated for a range of computer vision tasks (see Section~\ref{sec:relwork} for examples). Deep convolutional neural networks (CNNs)~\cite{Krizhevsky2012} represent the state-of-the-art method for a substantial number of visual tasks~\cite{he2016deep,long15fcn,ren2015faster}, which makes DA a critical issue for their practical application. The activations of such CNNs, and the interactions between them, are naturally represented as tensors, meaning that DA should also be applied using this representation. We show in Section~\ref{sec:ex} that the proposed method outperforms all comparators in DA of the convolutional activations of CNNs.

Vectorization also often results in the so-called \textit{curse of dimensionality}~\cite{shakhnarovich2011face}, as the matrices representing the relationships between vectorized tensors have $n^2$ elements, where $n$ is the number of elements in the tensor. This leads to errors in the estimation of this large number of parameters and high computational complexity. Furthermore, after vectorization, many existing approaches become sensitive to the scarcity of source data (compared to the number of dimensions) and noise in the labels.  The proposed direct tensor method uses much lower dimensional entities, thus avoiding these estimation problems.

\begin{figure}[!tb]
	\setlength{\abovecaptionskip}{10pt}
	\centering
	\includegraphics[width=2.8in,angle=0]{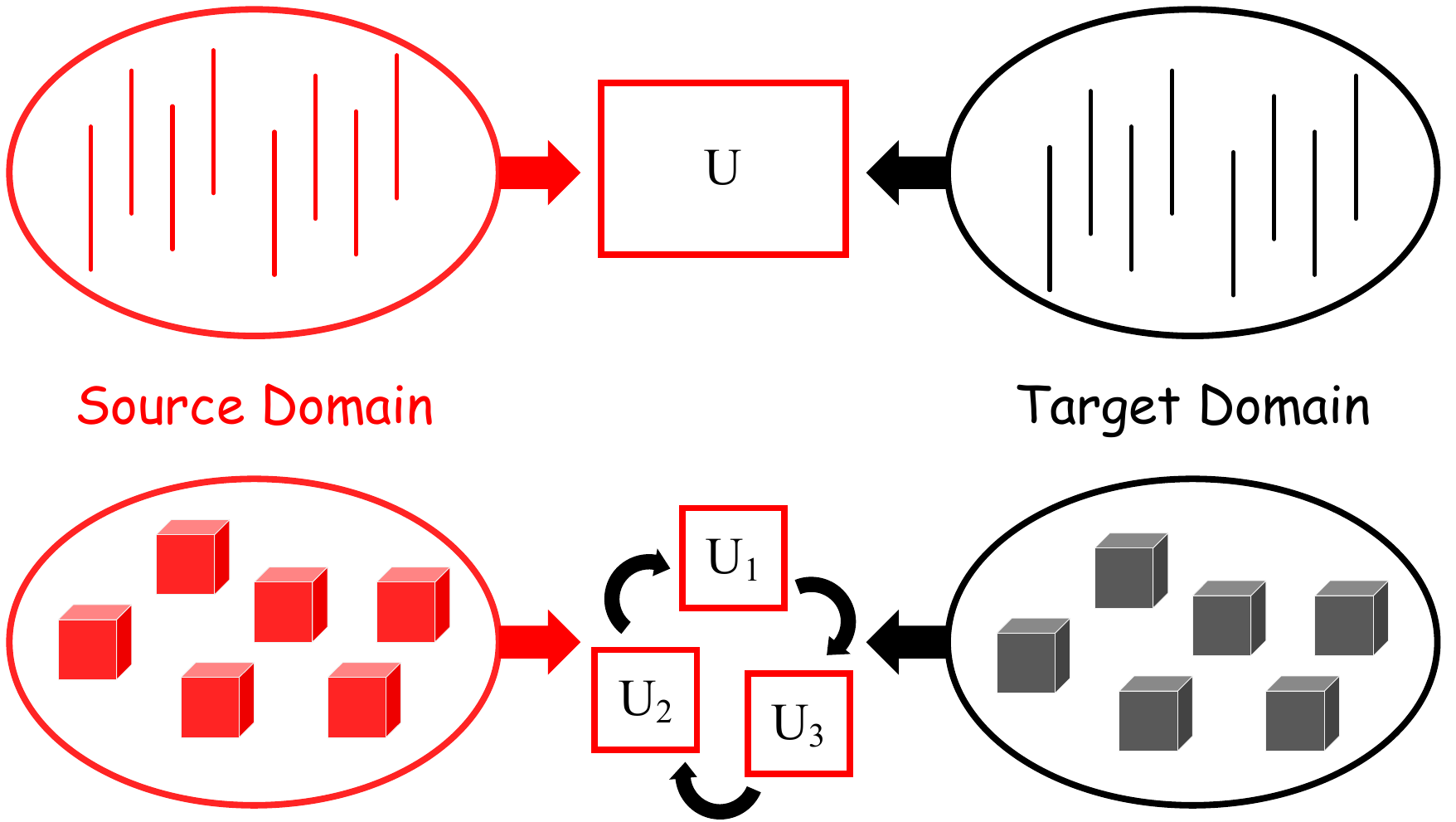}\vspace{-5pt}
	\caption{Vector subspace (top) vs. tensor subspace (bottom). Third-order (3-mode) tensors are used as an example. Compared to the vector subspace, the tensor subspace consists of a set of subspaces characterizing each mode respectively. Higher-order tensor modeling offers us an opportunity to investigate multiple interactions and couplings that capture the commonality and differences between domains.}
	\label{fig:tsubspace}
\end{figure}

To address these issues we propose to learn an invariant tensor subspace that is able to adapt the tensor representations directly. The key question is thus whether we can \textit{ find an invariant tensor subspace such that the domain discrepancy is reduced when the source data are adapted into the target domain}.
Following this idea, a novel approach termed Tensor-Aligned Invariant Subspace Learning (TAISL) is proposed for unsupervised DA. By introducing a set of alignment matrices, the tensor representations from the source domain are aligned to an underlying tensor subspace shared by the target domain. As illustrated in Fig.~\ref{fig:tsubspace}, the tensor subspace is able to preserve the intrinsic structure of representations by modeling the correlation between different modes. Instead of executing a holistic adaptation (where all feature dimensions would be taken into account), our approach performs mode-wise partial adaptation (where each mode is adapted separately) to avoid the curse of dimensionality. Seeking such a tensor subspace and learning the alignment matrices are consequently formulated into a joint optimization problem. We also propose an alternating minimization scheme, which allows the problem to be effectively optimized by off-the-shelf solvers.

Extensive experiments on cross-domain visual recognition demonstrate the following merits of our approach: i) it effectively reduces the domain discrepancy and preserves the discriminative power of the original representations; ii) it is applicable to small-sample-size adaptation, even when there is only one source sample per category; iii) it is robust to noisy labels; iv) it is computationally efficient, because the tensor subspace is constructed in a much smaller space than the vector-form paradigm; and v) it shows superior performance over state-of-the-art vector representation-based approaches in both the classification accuracy and computation time. Source code is made available online at: {\tt https://github.com/poppinace/TAISL}.

\section{Related work}
\label{sec:relwork}
Our work is closely related to subspace-based unsupervised DA and tensor representations.

\paragraph{Subspace-based domain adaptation.} Gopalan \etal \cite{gopalan2011domain} present one of the first visual DA approaches, which samples a finite set of subspaces along geodesic flows to bridge the source and target domains. Later in~\cite{gong12geodesic}, Gong \etal kernelize this idea by integrating an infinite number of subspaces that encapsulate the domain commonness and difference in a smooth and compact manner. Recently,~\cite{fernando13sa} argues that it is sufficient to directly align  the subspaces of two domains using a linear projection. Intuitively, such a linear mapping defines a shift of viewing angle that snapshots the source data from the target perspective. Subsequently,~\cite{aljundi2015landmarks} extends~\cite{fernando13sa} in a landmark-based kernelized paradigm. The performance improvement is due to the nonlinearity of the Gaussian kernel and sample reweighting. Alternatively,~\cite{shao2014ltsl} imposes a low-rank constraint during the subspace learning to reconstruct target samples with relevant source samples. More recently,~\cite{sun2016return} proposes to use the covariance matrix, a variant of the subspace, to characterize the domain, the adaptation is then cast as two simple but effective procedures of whitening the source data and recoloring the target covariance.

\paragraph{Tensor representations.} Tensor representations play a vital role in many computer vision applications~\cite{Kim09TCCA,Krizhevsky2012,li2007robust,vasilescu2002multilinear}. At the early stage of face representations,~\cite{vasilescu2002multilinear} introduced the idea of ``tensorfaces'' to jointly model multiple variations (viewpoint, expression, illumination, etc.).~\cite{li2007robust} achieves robust visual tracking by modeling frame-wise appearance using tensors.~\cite{Kim09TCCA} proposes tensor-based canonical correlation analysis as a representation for action recognition and detection. In other low-level tasks, such as image inpainting and image synthesis~\cite{Zhao15CP}, modeling images as a tensor is also a popular choice.

More recently, the most notable example is the deep CNNs~\cite{Krizhevsky2012}, as convolutional activations are intrinsically represented as tensors. The state-of-the-art performance of generic visual recognition and semantic image segmentation benefits from fully-convolutional models~\cite{he2016deep,long15fcn}. Aside from this, by reusing convolutional feature maps, proposal generation and object detection can be performed simultaneously in a faster R-CNN fashion~\cite{ren2015faster}. Yet, convolutional activations still suffer from the domain shift~\cite{lu20172dsa,yosinski2014transferable}. How to adapt convolutional activations effectively remains an open question.

Tensor representations are important, while solutions to adapt them are limited. To fill this gap, we present one of the first DA approaches for tensor representations adaptation.

\section{Learning an invariant tensor subspace}

Before we present our technical details, some mathematical background related to tensor decomposition is provided. In the following mathematical expressions, we denote matrices and tensors by uppercase boldface letters and calligraphic letters, respectively, such as $\mat{U}$ and $\mathcal{U}$.

\subsection{Tensor decomposition revisited}
\label{subsec:tucker}

A tensor of order (mode) $K$ is denoted by $\mathcal{X}\in{\mathbb{R}^{n_1 \times ... \times n_K}}$. Its \texttt{mode-$k$ product} is defined as \mbox{$\mathcal{X}\times_k\mat{V}$}.
The operator $\times_k$ indicates matrix multiplication performed along the $k$-th mode. Equivalently, \mbox{$(\mathcal{X}\times_k\mat{V})_{(k)}=\mat{V}\mat{X}_{(k)}$}, where $\mat{X}_{(k)}$ is called the \texttt{mode-$k$ matrix unfolding}, a procedure of reshaping a tensor $\mathcal{X}$ into a matrix $\mat{X}_{(k)}\in{\mathbb{R}^{n_k\times n_1...n_{k-1}n_{k+1} ... n_K}}$.

In this paper we draw upon \textit{Tucker Decomposition}~\cite{kolda2009tensor} to generate tensor subspaces. Tucker decomposition decomposes a $K$-mode tensor $\mathcal{X}$ into a core tensor multiplied by a set of factor matrices along each mode as follows:
\begin{equation}\label{eq:tucker}
\small
\setlength\abovedisplayskip{1pt}
\setlength\belowdisplayskip{1pt}
\begin{aligned}
\mathcal{X} = \mathcal{G}\times_1 \mat{U}^{(1)} \times_2 \mat{U}^{(2)} \times_3 \cdots \times_K \mat{U}^{(K)} = \ldbracket\mathcal{G}; \mathcal{U}\rdbracket {\kern 4pt}
\end{aligned}\\,
\end{equation}
where $\mathcal{G}\in{\mathbb{R}^{d_1\times ... \times d_K}}$ is the core tensor, and $\mat{U}^{(k)} \in{\mathbb{R}^{n_k \times d_k}}$ denotes the factor matrix of the $k$-th mode. The column space of $\mat{U}^{(k)}$ expands the corresponding signal subspace. To simply the notation, with $\mathcal{U} = \{\mat{U}^{(k)}\}_{k=1,...,K}$, Tucker decomposition can be concisely represented as the right part of Eq.~\ref{eq:tucker}. Here, $\mathcal{U}$ is the tensor subspace, and $\mathcal{G}$ is the tensor subspace representation of $\mathcal{X}$. Alternatively, via the Kronecker product, Tucker decomposition can be expressed in matrix form as \mbox{$\mat{X}_{(k)}=\mat{U}^{(k)}\mat{G}_{(k)}\mat{U}_{\setminus k}^T$}, where
\begin{equation}\label{eq:tucker_minus}
\small
\setlength\abovedisplayskip{1pt}
\setlength\belowdisplayskip{1pt}
\begin{aligned}
\mat{U}_{\setminus k}=\mat{U}^{(K)} \otimes \cdots \otimes \mat{U}^{(k+1)} \otimes \mat{U}^{(k-1)} \otimes \cdots \otimes \mat{U}^{(1)}
\end{aligned}\\,
\end{equation}
and $\otimes$ denotes the Kronecker product.

\subsection{Naive tensor subspace learning}
\label{subsec:tsl}

Perhaps the most straight-forward way to adapt domains is to assume an invariant subspace between the source domain $S$ and the target domain $T$. This assumption is reasonable when the domain discrepancy is not very large. With this idea, we first introduce the Naive Tensor Subspace Learning (NTSL) below, which can be viewed as a baseline of our approach.

Given $N_s$ samples $\{\mathcal{X}^{n}_{s}\}_{n=1,...,N_s}$ from source domain, each sample is denoted as a $K$-mode tensor $\mathcal{X}^{n}_{s}\in\mathbb{R}^{n_1\times...\times n_K}$. For simplicity, $N_s$ samples are stacked into a $(K+1)$-mode tensor $\mathcal{X}_s \in{\mathbb{R}^{n_1 \times ... \times n_K \times N_s}}$. Similarly, let $\mathcal{X}_t \in{\mathbb{R}^{m_1 \times ... \times m_K \times N_t}}$ be a set of $N_t$ samples from the target domain $T$. In general, we consider $n_k=m_k, k=1,2,...,K$, because the case with heterogeneous data is out the scope of this paper. Provided that $S$ and $T$ share a underlying tensor subspace $\mathcal{U} = \{\mat{U}^{(k)}\}_{k=1,...,K}, \mat{U}^{(k)}\in\mathbb{R}^{n_k\times d_k}$, on the basis of Tucker decomposition, seeking $\mathcal{U}$ is equivalent to solve the following optimization problem as
\begin{equation}\label{eq:tsl_obj}
\small
\setlength\abovedisplayskip{1pt}
\setlength\belowdisplayskip{1pt}
\begin{aligned}
\min\limits_{\mathcal{U}, \mathcal{G}_s, \mathcal{G}_t} & \|\mathcal{X}_s - \ldbracket\mathcal{G}_s; \mathcal{U}\rdbracket\|^2_F + \|\mathcal{X}_t - \ldbracket\mathcal{G}_t; \mathcal{U}\rdbracket\|^2_F {\kern 4pt}\\
{\rm{s.t.}} & {\kern 4pt} \forall k, {\kern 2pt} {\mat{U}}^{(k)T}{\mat{U}^{(k)}} = \mat{I}
\end{aligned}\\,
\end{equation}
where $\mathcal{G}_s$ and $\mathcal{G}_t$ denote the tensor subspace representation of $\mathcal{X}_s$ and $\mathcal{X}_t$, respectively. $\mat{I}$ is an identity matrix with appropriate size. Here $\mathcal{U}$ is the invariant tensor subspace in which the idea of DA lies. One can employ the off-the-shelf Tucker decomposition algorithm to solve Eq.~\eqref{eq:tsl_obj} effectively. Once the optimum $\mathcal{U}^*$ is identified, $\mathcal{G}_s$ can be obtained by the following straight-forward multilinear product as
\begin{equation}\label{eq:proj}
\small
\setlength\abovedisplayskip{1pt}
\setlength\belowdisplayskip{1pt}
\begin{aligned}
\mathcal{G}}_s =\mathcal{X}_s \times_1 \mat{U^*}^{(1)T} \times_2 \mat{U^*}^{(2)T} \times_3 \cdots \times_K\mat{U^*}^{(K)T
\end{aligned}\\.
\end{equation}
A similar procedure can be applied to derive $\mathcal{G}_t$. Next, if DA is evaluated in the context of classification, one can learn a linear classifier with $\mathcal{G}_s$ and source label $L_s$, and then verifies the classification performance on $\mathcal{G}_t$.

\subsection{Tensor-aligned invariant subspace learning}
\label{subsec:tasl}

Eq.~\eqref{eq:tsl_obj} assumes a shared subspace between two domains. However, when the domain discrepancy becomes larger, enforcing only a shared subspace is typically not sufficient. To address this, we present Tensor-Aligned Invariant Subspace Learning (TAISL) which aims to reduce the domain discrepancy more explicitly. Motivated by the idea that a simple linear transformation can effectively reduce the domain discrepancy~\cite{baktashmotlagh2013dip,fernando13sa}, we introduce a set of alignment matrices into Eq.~\eqref{eq:tsl_obj}. This yields the following optimization problem as
\begin{equation}\label{eq:tasl_naive}
\small
\setlength\abovedisplayskip{1pt}
\setlength\belowdisplayskip{1pt}
\begin{aligned}
\min\limits_{\mathcal{U}, \mathcal{G}_s, \mathcal{G}_t, \mathcal{M}} & \|\ldbracket \mathcal{X}_s; \mathcal{M}\rdbracket - \ldbracket\mathcal{G}_s; \mathcal{U}\rdbracket\|^2_F + \|\mathcal{X}_t - \ldbracket\mathcal{G}_t; \mathcal{U}\rdbracket\|^2_F {\kern 4pt}\\
{\rm{s.t.}} & {\kern 4pt} \forall k, {\kern 2pt} {\mat{U}}^{(k)T}{\mat{U}^{(k)}} = \mat{I}
\end{aligned}\\,
\end{equation}
where $\mathcal{M} = \{\mat{M}^{(k)}\}_{k=1,...,K}$, $\mat{M}^{(k)} \in{\mathbb{R}^{m_k \times n_k}}$. With $\mathcal{M}$, samples from $S$ can be linearly aligned to $T$. Here, $\mat{M}^{(k)}$ is unconstrained, which is undesirable in a well-defined optimization problem. To narrow down the search space, a natural choice to regularize $\mat{M}^{(k)}$ is the Frobenius norm $\|\mat{M}^{(k)}\|_F^2$. However,~\cite{Sinno11TCA} suggests that the original data variance should be preserved after the alignment. Otherwise, there is a high probability the projected data will cluster into a single point. As a consequence, we employ a PCA-like constraint on $\mathcal{M}$ to maximally preserve the data variance. This gives our overall optimization problem
\begin{equation}\label{eq:tasl_obj}
\small
\setlength\abovedisplayskip{1pt}
\setlength\belowdisplayskip{1pt}
\begin{aligned}
\min\limits_{\mathcal{U}, \mathcal{G}_s, \mathcal{G}_t, \mathcal{M}} & \|\ldbracket \mathcal{X}_s; \mathcal{M}\rdbracket - \ldbracket\mathcal{G}_s; \mathcal{U}\rdbracket\|^2_F + \|\mathcal{X}_t - \ldbracket\mathcal{G}_t; \mathcal{U}\rdbracket\|^2_F\\
& {\kern 50pt} + \lambda \|\ldbracket\ldbracket \mathcal{X}_s; \mathcal{M}\rdbracket; \mathcal{M}^T\rdbracket - \mathcal{X}_s\|^2_F {\kern 4pt}\\
{\rm{s.t.}} & {\kern 4pt} \forall k, {\kern 2pt} {\mat{U}}^{(k)T}{\mat{U}^{(k)}} = \mat{I}, {\mat{M}^{(k)}}{\mat{M}^{(k)T}} = \mat{I}
\end{aligned}\\,
\end{equation}
where $\lambda$ is a weight on the regularization term. Intuitively, the regularization term measures how well $\mathcal{M}$ reconstructs the source data. Note that, in contrast $\mat{U}^{(k)}$, which is column-wise orthogonal, $\mat{M}^{(k)}$ is row-wise orthogonal. Moreover, both $\mat{U}^{(k)}$ and $\mat{M}^{(k)}$ have no effect on the $(K+1)$-th mode, because the adaptation of data dimension makes no sense.

\paragraph{\textit{Relation to subspace alignment.}} As mentioned in Section~\ref{sec:relwork}, the seminal subspace alignment (SA) framework is introduced in~\cite{fernando13sa}. Given two vector subspaces $\mat{U}_s$ and $\mat{U}_t$ of two domains, the domain discrepancy is measured by the Bregman divergence as $\|\mat{U}_s\mat{M}-\mat{U}_t\|_F^2$. Here $\mat{M}$ aligns the subspaces. In our formulation, $\mathcal{M}$ seems to align the data directly at the first glance. However, if one takes the properties of the mode-$k$ product into account, one can see that this is not the case. According to the definition of the Tucker decomposition, for $\mathcal{X}_s$, we have $\mathcal{X}_s = \mathcal{G}_s\times_1 \mat{U}_s^{(1)} \times_2 \cdots \times_K \mat{U}_s^{(K)}$, so $\ldbracket \mathcal{X}_s; \mathcal{M}\rdbracket$ can be expanded as
\begin{equation}\label{eq:tucker_xsm}
\small
\setlength\abovedisplayskip{1pt}
\setlength\belowdisplayskip{1pt}
\begin{aligned}
&\mathcal{X}_s \times_1 \mat{M}^{(1)} \times_2 \cdots \times_K \mat{M}^{(K)}\\
= {\kern 4pt} &\mathcal{G}_s\times_1 (\mat{M}^{(1)}\mat{U}_s^{(1)}) \times_2 \cdots \times_K (\mat{M}^{(K)}\mat{U}_s^{(K)})
\end{aligned}\\.
\end{equation}
That is, the alignment of the tensor is equivalent to the alignment of the tensor subspace. As a consequence, our approach can be viewed as a natural generalization of~\cite{fernando13sa} to the multidimensional case. However, unlike SA, in which the DA and subspaces are learned separately, the alignment matrices $\mathcal{M}$ and the tensor subspace $\mathcal{U}$ in our approach are learned jointly in an unified paradigm.

\section{Optimization}
\label{sec:optimize}

Here we discuss how to solve the problem in Eq.~\eqref{eq:tasl_obj}. Since $\mathcal{M}$ and $\mathcal{U}$ are coupled in Eq.~\eqref{eq:tasl_obj}, it is hard for a joint optimization. A general strategy is to use alternative minimization to decompose the problem into subproblems and to iteratively optimize these subproblems until convergence, acquiring an approximate solution~\cite{shao2014ltsl,zhang2016exploring,zhang2016dictionary}.

\paragraph{\textit{Optimize $\mathcal{U}$, $\mathcal{G}_s$, and $\mathcal{G}_t$ given $\mathcal{M}$}:} By introducing an auxiliary variable $\mathcal{Z} = \ldbracket \mathcal{X}_s; \mathcal{M}\rdbracket$, the subproblem over $\mathcal{U}$, $\mathcal{G}_s$ and $\mathcal{G}_t$ can be given as
\begin{equation}\label{eq:tasl_u}
\small\
\setlength\abovedisplayskip{1pt}
\setlength\belowdisplayskip{1pt}
\begin{aligned}
\min\limits_{\mathcal{U}, \mathcal{G}_s, \mathcal{G}_t} & \|\mathcal{Z} - \ldbracket\mathcal{G}_s; \mathcal{U}\rdbracket\|^2_F + \|\mathcal{X}_t - \ldbracket\mathcal{G}_t; \mathcal{U}\rdbracket\|^2_F {\kern 4pt}\\
{\rm{s.t.}} & {\kern 4pt} \forall k, {\kern 2pt} {\mat{U}}^{(k)T}{\mat{U}^{(k)}} = \mat{I}
\end{aligned}\\,
\end{equation}
which is exactly the same problem in Eq.~\eqref{eq:tsl_obj} and can be easily solved in the same paradigm.

\paragraph{\textit{Optimize $\mathcal{M}$ given $\mathcal{U}$, $\mathcal{G}_s$, and $\mathcal{G}_t$}:} By introducing another auxiliary variable $\mathcal{Y} = \ldbracket\mathcal{G}_s; \mathcal{U}\rdbracket\in{\mathbb{R}^{n_1\times\cdots\times n_K\times N_s}}$, we arrive at the subproblem over $\mathcal{M}$ as
\begin{equation}\label{eq:tasl_m}
\small
\setlength\abovedisplayskip{1pt}
\setlength\belowdisplayskip{1pt}
\begin{aligned}
\min_{\mathcal{M}} & {\kern 4pt} \|\ldbracket \mathcal{X}_s; \mathcal{M}\rdbracket - \mathcal{Y}\|^2_F + \lambda \|\ldbracket\ldbracket \mathcal{X}_s; \mathcal{M}\rdbracket; \mathcal{M}^T\rdbracket - \mathcal{X}_s\|^2_F {\kern 4pt}\\
{\rm{s.t.}} & {\kern 4pt} \forall k, {\kern 2pt} {\mat{M}^{(k)}}{\mat{M}^{(k)T}} = \mat{I}
\end{aligned}\\.
\end{equation}
Directly solving $\mathcal{M}$ is intractable, but we can optimize each $\mat{M}^{(k)}$ individually. To this end, Eq.~\eqref{eq:tasl_m} needs to be reformulated further. Let $\mat{Y}_{(k)}$ be the $k$-mode unfolding matrix of tensor $\mathcal{Y}$, and $\mat{M}_{\setminus k} = \mat{I}\otimes\mat{M}^{(K)}\otimes\cdots\otimes\mat{M}^{(k+1)}\otimes\mat{M}^{(k-1)}\otimes\cdots\otimes\mat{M}^{(1)}$. Unfolding the $k$-th mode of the first term in Eq.~\eqref{eq:tasl_m} can be given as
\begin{equation}\label{eq:tasl_m1}
\small
\setlength\abovedisplayskip{1pt}
\setlength\belowdisplayskip{1pt}
\begin{aligned}
\|\left[\ldbracket \mathcal{X}_s; \mathcal{M}\rdbracket - \mathcal{Y}\right]_{(k)}\|^2_F = \|\mat{M}^{(k)}\mat{X}_{s(k)}\mat{M}^T_{\setminus k} - \mat{Y}_{(k)}\|^2_F
\end{aligned}\\.
\end{equation}
For the regularizer, since $\mathcal{M}$ cannot be directly decomposed into individual $\mat{M}^{(k)}$, we raise an assumption here to make the optimization tractable in practice. Considering that
\begin{equation}\label{eq:tasl_m2}
\small
\setlength\abovedisplayskip{1pt}
\setlength\belowdisplayskip{1pt}
\begin{aligned}
{\ldbracket}{\ldbracket} \mathcal{X}_s; \mathcal{M}\rdbracket; \mathcal{M}^T\rdbracket = \mathcal{X}_s \times_1 (\mat{M}^{(1)T}\mat{M}^{(1)}) \times_2 ... {\kern 4pt}\\
\times_K (\mat{M}^{(K)T}\mat{M}^{(K)}) {\kern 4pt}
\end{aligned}\\,
\end{equation}
for the $k$-th mode, we have
\begin{equation}\label{eq:tasl_m3}
\small
\setlength\abovedisplayskip{1pt}
\setlength\belowdisplayskip{1pt}
\begin{aligned}
\left[\mathcal{X}_s \times_k (\mat{M}^{(k)T}\mat{M}^{(k)})\right]_{(k)} = \mat{M}^{(k)T}\mat{M}^{(k)}\mat{X}_{s(k)} {\kern 4pt}
\end{aligned}\\.
\end{equation}
Provided that ${\mbox{\boldmath{$M$}}}^T_{\setminus k}$ is given and all $\mat{M}^{(i)}$s for $i\neq k$ well preserve the energy of $\mathcal{X}_{s}$, i.e., we assume $\mat{M}^{(i)T}\mat{M}^{(i)}\approx \mat{I}$, \mbox{$i\neq k$}. Though this assumption seems somewhat heuristic, we show later in experiments the loss decreases normally, which suggests it is at least a good approximation. Hence, optimizing Eq.~\eqref{eq:tasl_m} over $\mathcal{M}$ can be decomposed to $K$ subproblems. The $k$-th subproblem over $\mat{M}^{(k)}$ gives
\begin{equation}\label{eq:tasl_m4}
\small
\setlength\abovedisplayskip{2pt}
\setlength\belowdisplayskip{2pt}
\begin{aligned}
\min_{{\mat{\scriptstyle M^{(k)}}}} & \|\mat{M}^{(k)}\mat{Q}_{(k)} {\kern -1pt}-{\kern -1pt} \mat{Y}_{(k)}\|^2_F {\kern -1pt}+{\kern -1pt} \lambda \|\mat{M}^{(k)T}{\kern -2pt}\mat{M}^{(k)}{\kern -1pt}\mat{X}_{s(k)}{\kern -2pt} -{\kern -1pt} \mat{X}_{s(k)}\|^2_F\\
{\rm{s.t.}} & {\kern 4pt} \forall k, {\kern 2pt} {\mat{M}^{(k)}}{\mat{M}^{(k)T}} = \mat{I}
\end{aligned}\\,
\end{equation}
where $\mat{Q}_{(k)} = \mat{X}_{s(k)}\mat{M}^T_{\setminus k}$. Notice that
\begin{equation}\label{eq:tasl_m5}
\small
\setlength\abovedisplayskip{2pt}
\setlength\belowdisplayskip{2pt}
\begin{aligned}
& \|\mat{M}^{(k)T}{\kern -2pt}\mat{M}^{(k)}{\kern -2pt}\mat{X}_{s(k)}{\kern -1pt}-{\kern -1pt}\mat{X}_{s(k)}\|^2_F = \|\mat{X}_{s(k)}\|_F^2 - \|\mat{M}^{(k)}{\kern -2pt}\mat{X}_{s(k)}\|_F^2.
\end{aligned}
\end{equation}
Since $\|\mat{X}_{s(k)}\|_F^2$ remains unchanged during the optimization of $\mat{M}^{(k)}$, this term can be ignored. Therefore, Eq.~\eqref{eq:tasl_m4} can be further simplified as
\begin{equation}\label{eq:tasl_m6}
\small
\setlength\abovedisplayskip{1pt}
\setlength\belowdisplayskip{1pt}
\begin{aligned}
\min_{{\mat{\scriptstyle M}^{(k)}}} & {\kern 4pt} \|\mat{M}^{(k)}\mat{Q}_{(k)} - \mat{Y}_{(k)}\|^2_F - \lambda \|\mat{M}^{(k)}\mat{X}_{s(k)}\|_F^2 {\kern 4pt}\\
{\rm{s.t.}} & {\kern 4pt} \forall k, {\kern 2pt} {\mat{M}^{(k)}}{\mat{M}^{(k)T}} = \mat{I}
\end{aligned}\\.
\end{equation}
Finally, by replacing $\mat{P} = \mat{M}^{(k)T}$, we can transform Eq.~\eqref{eq:tasl_m6} into a standard orthogonality constraint based optimization problem as
\begin{equation}\label{eq:tasl_m7}
\small
\setlength\abovedisplayskip{1pt}
\setlength\belowdisplayskip{1pt}
\begin{aligned}
\min_{{\mat{\scriptstyle P}}} & {\kern 4pt} \|\mat{Q}_{(k)}^T\mat{P} - \mat{Y}_{(k)}^T\|^2_F - \lambda \|\mat{X}_{s(k)}^T\mat{P}\|_F^2 {\kern 4pt}\\
{\rm{s.t.}} & {\kern 4pt} \forall k, {\kern 2pt} {\mat{P}^T}{\mat{P}} = \mat{I}
\end{aligned}\\,
\end{equation}
which can be effectively solved by a standard solver, like the solver presented in~\cite{wen2013feasible}. This alternating minimization approach is summarized in Algorithm~\ref{alg:am}. We observe that the optimization converges only after several iterations.

\section{Results and discussion}
\label{sec:ex}

In this section, we first illustrate the merits of our approach on a standard DA dataset, and then focus on comparisons with related and state-of-the-art methods.

\subsection{Datasets, protocol, and baselines}
\label{sec:data_protocol}

\paragraph{Office--Caltech10 (OC10) dataset.} OC10 dataset~\cite{gong12geodesic} is the extension of Office~\cite{saenko2010adapting} dataset by adding another \textit{Caltech} domain, resulting in four domains of \textit{Amazon} (A), \textit{DSLR} (D), \textit{web-cam} (W), and \textit{Caltech} (C). 10 common categories are chosen, leading to around 2500 images and 12 DA problem settings. This dataset reflects the domain shift caused by appearance, viewpoint, background and image resolution. For short, a DA task is denoted by S$\rightarrow$T.

\begin{algorithm}[!t]
	\small
	\caption{Alternating minimization for TAISL}
	\label{alg:am}
	\KwIn{Source data: $\mathcal{X}_s$; Target data: $\mathcal{X}_t$\\}
	\KwOut{Tensor subspace: $\mathcal{U}$; Alignment matrices: $\mathcal{M}$}
	\textbf{Initialize:} $\mat{M}^{(k)}=I, k=1,...,K$;\\Tensor subspace dimensionality: $d_k, k=1,...,K$;\\Weight coefficient: $\lambda$;\\ Maximum iteration: $T$;\\
	\For{$t \gets 1$ \textbf{to} $T$} {
		Subspace learning over $\{\mathcal{U}$, $\mathcal{G}_s$, $\mathcal{G}_t\}$ as per Eq.~\eqref{eq:tasl_u};\\
		\For{$k \gets 1$ \textbf{to} $K$} {
			Optimization over $\mat{M}^{(k)}$ as per Eq.~\eqref{eq:tasl_m7};\\
		}
		Check for convergence;\\
	}
\end{algorithm}

\vspace{-12.5pt}
\paragraph{ImageNet--VOC2007 (IV) dataset.} We also evaluate our method on the widely-used ImageNet~\cite{Deng2009} and PASCAL VOC2007~\cite{everingham2010pascal} datasets. The same 20 categories of the VOC2007 are chosen from the ImageNet 2012 dataset to form the source domain, and the \texttt{test} set of VOC2007 is used as the target domain. Notice that VOC 2007 is a multi-label dataset. IV dataset reflects the shift when transferring from salient objects to objects in complex scenes. We use this to verify the effectiveness of DA approaches when multiple labels occur.

\vspace{-12.5pt}
\paragraph{Experimental protocol.} In this paper, we focus on the small-sample-size adaptation, because if enough source and target data are made available, we have better choices with deep adaptation techniques~\cite{ganin2015unsupervised,sener2016learning} to co-adapt the feature representation, domain distributions and the classifier. In particular, the sampling protocol in~\cite{gong12geodesic} is used. Concretely, for both datasets, 20 images are randomly sampled from each category of the source domain (8 images if the domain is \textit{web-cam} or \textit{DSLR}) in each trials. The mean and standard deviation of average multi-class accuracy over 20 trials are reported on OC10 dataset. For the IV dataset, we follow the standard evaluation criterion~\cite{everingham2010pascal} to use the average precision (AP) as the measure. Similarly, the mean and standard deviation of AP over 10 trials are reported for each category.

\vspace{-12.5pt}
\paragraph{Baseline approaches.} Several approaches are employed for comparisons:
\squishlist
	\item \textit{No Adaptation (NA)}: NA indicates to train a classifier directly using the labeled source data and applies to the target domain. This is a basic baseline.
	\item \textit{Principal Component Analysis (PCA)}: PCA is a direct baseline compared to our NTSL approach. It assumes an invariant vector subspace between domains.
	\item \textit{Daum\'{e} III}~\cite{daume2007frustratingly}: Daum\'{e} III is a classical DA approach through augmenting the feature representations. Each source data point $\mathbf{x_s}$ is augmented to $\mathbf{x_s}'=(\mathbf{x_s}, \mathbf{x_s}, \mathbf{0})$, and each target data point $\mathbf{x_t}$ to $\mathbf{x_t}'=(\mathbf{x_t}, \mathbf{0}, \mathbf{x_t}).$
	\item \textit{Transfer Component Analysis (TCA)}~\cite{Sinno11TCA}: TCA formulates DA in a reproducing kernel Hilbert space by minimizing the maximum mean discrepancy measure.
	\item \textit{Geodesic Flow Kernel (GFK)}~\cite{gong12geodesic}: GFK proposes a closed-form solution to bridge the subspaces of two domains using a geodesic flow in a Grassmann manifold.
	\item \textit{Domain Invariant Projection (DIP)}~\cite{baktashmotlagh2013dip}: DIP seeks domain-invariant representations by matching the source and target distributions in a low-dimensional reproducing kernel Hilbert space.
	\item \textit{Subspace Alignment (SA)}~\cite{fernando13sa}: SA directly adopts a linear projection to match the differences between the source and target subspaces. Our approach is closely related to this method.
	\item \textit{Low-rank Transfer Subspace Learning (LTSL)}~\cite{shao2014ltsl}: LTSL imposes a low-rank constraint during the subspace learning to enforce only relevant source data are used to reconstruct the target data.
	\item \textit{Landmarks Selection Subspace Alignment (LSSA)}~\cite{aljundi2015landmarks}: LSSA extends SA via selecting landmarks and using further nonlinearity with Gaussian kernel.
	\item \textit{Correlation Alignment (CORAL)}~\cite{sun2016return}: CORAL characterizes domains using their covariance matrices. DA is performed through simple whitening and recoloring.
\squishend

Notice that, for a fair comparison, some methods, e.g., STM~\cite{Chu2013STM}, that take source labels into account during the optimization are not chosen for comparison, because TAISL does not utilize the information of source labels during DA.

\paragraph{Parameters setting.} We extract the convolutional activations from the CONV5\_3 layer of VGG--16 model~\cite{Simonyan14verydeep} as the tensor representation. We allow the input image to be of arbitrary size, so a simple spatial pooling~\cite{He2015} procedure is applied as the normalization. Specifically, each image will be mapped into a $6\times 6 \times 512$ third-order tensor. For those conventional approaches, convolutional activations are vectorized into a long vector as the representation. For NTSL and TAISL, we empirically set the tensor subspace dimensionality as $d_1=d_2=6$, and $d_3=128$. The first and second modes refer to the spatial location, and the third mode corresponds to the feature. We set such parameters with a motivation to preserve the spatial information and to seek the underlying commonness in the low-dimensional subspace. The weight parameter is set to $\lambda=1e^{-5}$, and the maximum iteration $T=10$. Note that we adopt these hyper parameters for all DA tasks when reporting the results. For the comparator approaches, parameters are set according to the suggestions of corresponding papers. One-vs-rest linear SVMs are used as the classifiers, and the penalty parameter $C_{svm}$ is fixed to $1$. Please refer to the Supplementary Materials for further details and results.

\subsection{Evaluation on the Office--Caltech10 dataset}
\label{subsec:eval_oc10}

Before we present the full DA results, we first highlight the merits of tensor subspaces for DA from three aspects: 1) quantifying the domain discrepancy to show how well TAISL preserves the discriminative power of the source domain, 2) evaluating the classification performance with a limited number of source/target data to see what scenarios TAISL could be applied in, and 3) replacing source data with noisy labels to verify whether TAISL can resist noise interference.

\begin{figure}[!tb]
	\centering
	\subfloat[]{\includegraphics[width=0.33\linewidth]{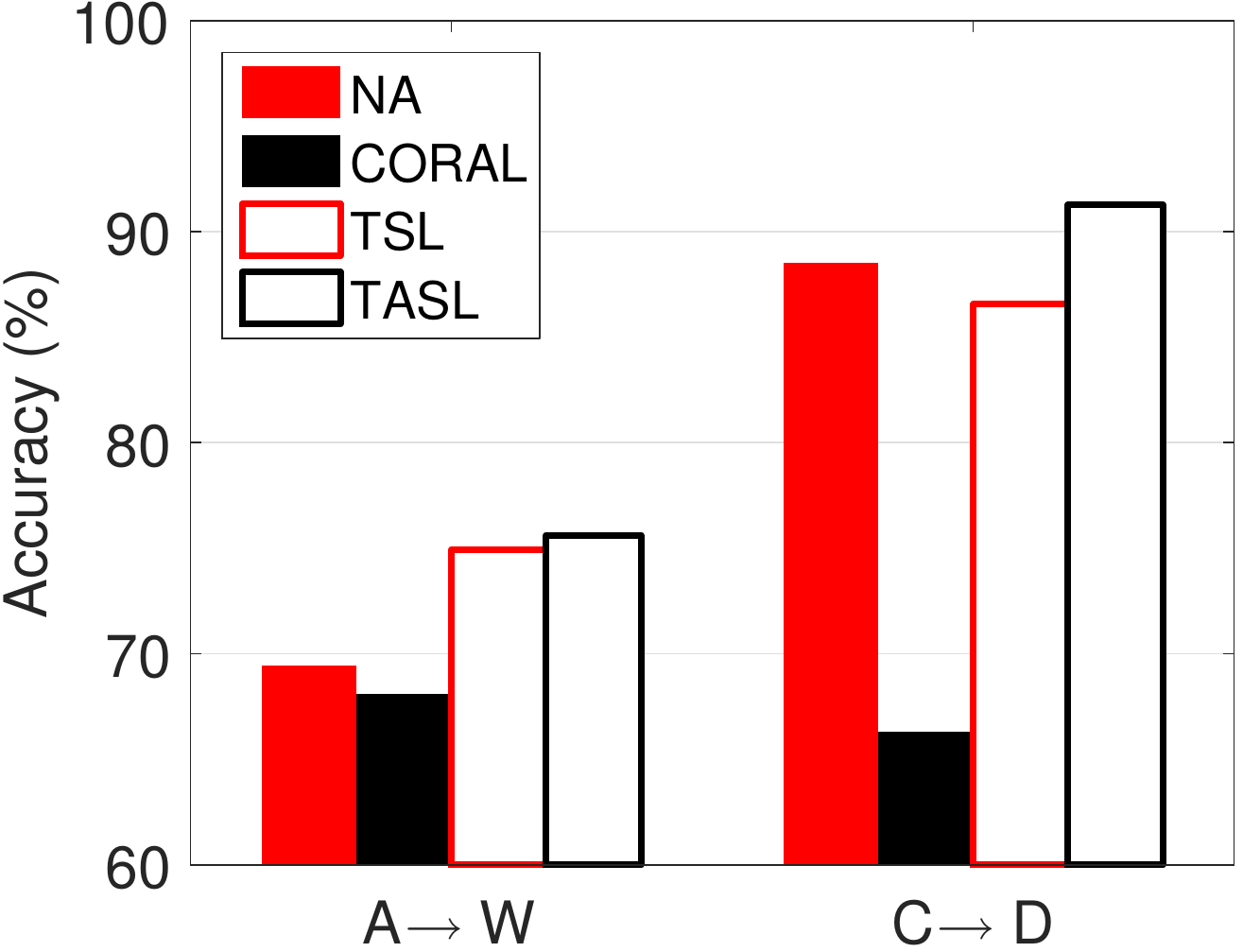}}
	\subfloat[]{\includegraphics[width=0.33\linewidth]{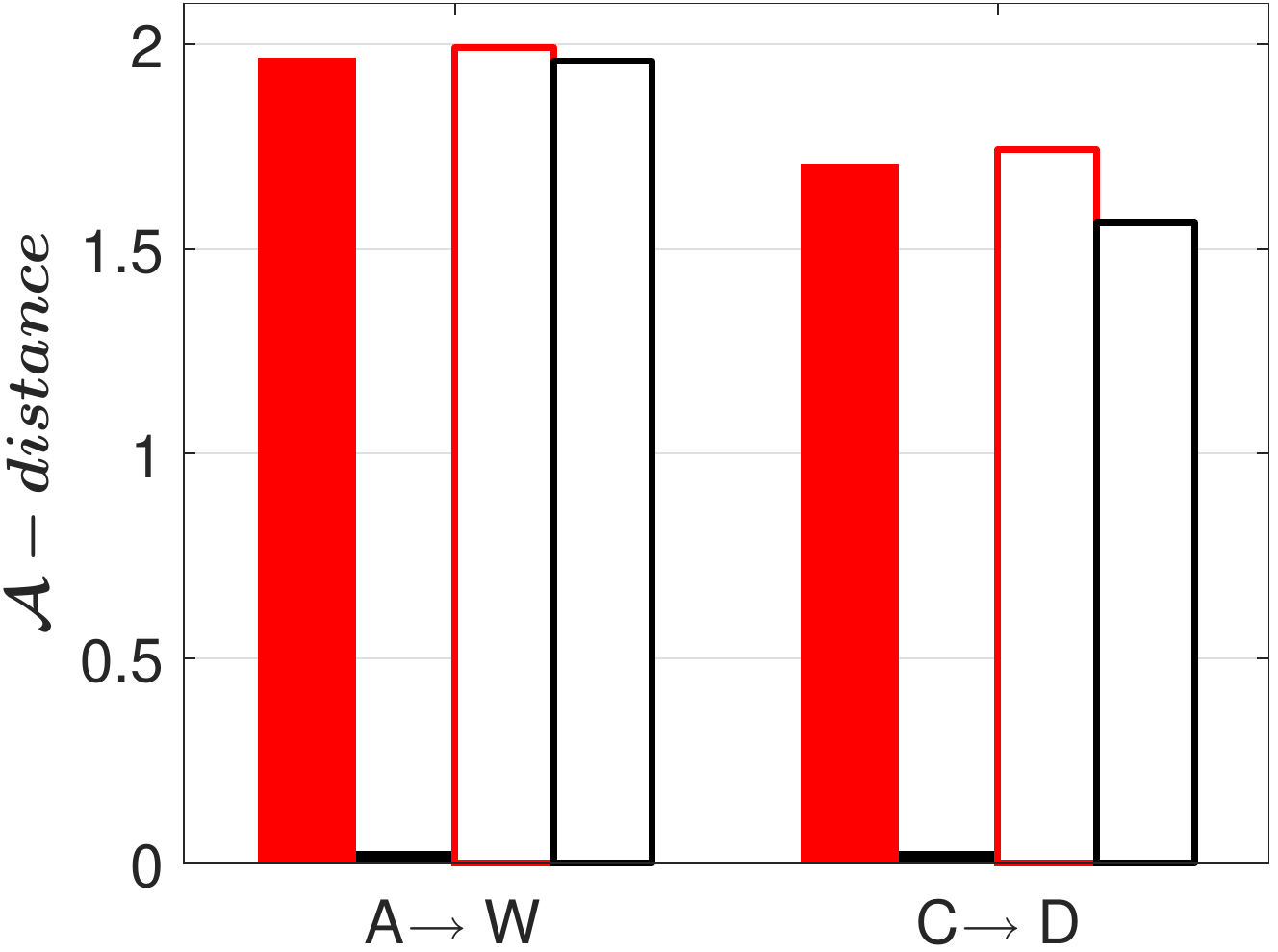}}
	\subfloat[]{\includegraphics[width=0.33\linewidth]{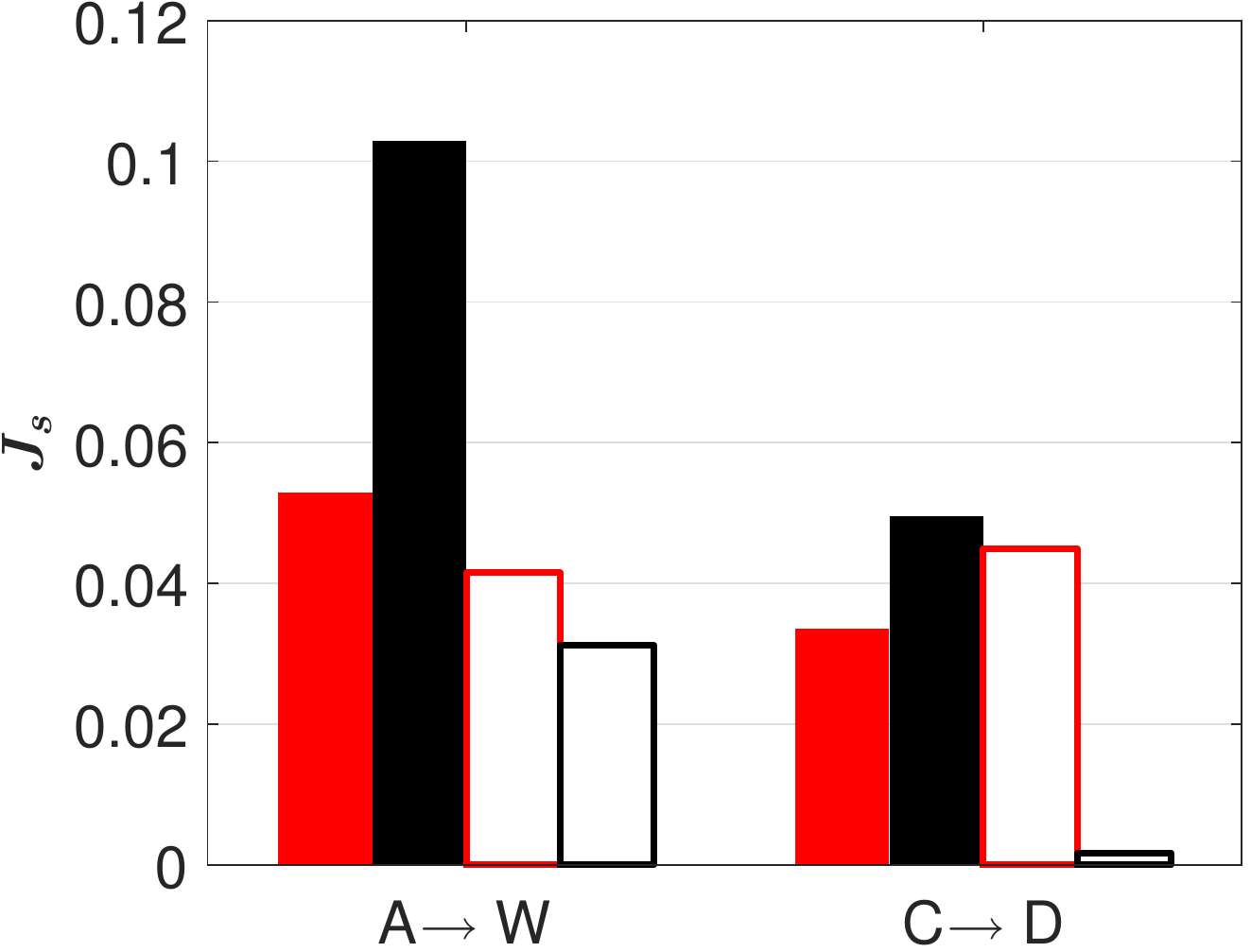}}
	\vspace{-10pt}
	\caption{Classification accuracy (a) and domain discrepancy measured by domain-level $\mathcal{A}$-distance (b) and class-level $J_s$ divergence (c) over two DA tasks.}
	\label{fig:A_dist}
\end{figure}

\paragraph{Quantifying the class-level domain discrepancy.} $\mathcal{A}$-distance has been introduced in~\cite{ben2007analysis} as a popular measure of domain discrepancy over two distributions. Estimating this distance involves pseudo-labeling the source domain $\mathcal{P}_s$ and target domain $\mathcal{P}_t$ as a binary classification problem. By learning a linear classifier, $\mathcal{A}$-distance can be estimated as $d_\mathcal{A}(\mathcal{P}_s,\mathcal{P}_t)=2(1-2\epsilon)$, where $\epsilon$ is the generalization error of the linear classifier. The lower $\mathcal{A}$-distance is, the better two distributions align, and vice versa. Given this measure, we empirically examine the correlation between the classification accuracy and $\mathcal{A}$-distance. Fig.~\ref{fig:A_dist}(a) and Fig.~\ref{fig:A_dist}(b) illustrate these two measures of several approaches on two DA tasks. Surprisingly, two measures exhibit a totally adverse tendency. The lowest classification accuracy conversely corresponds to the lowest $\mathcal{A}$-distance. As a consequence, at least for convolutional activations, we consider that \textit{the classification accuracy has low correlations with the domain-level discrepancy}. In an effort to explain such a phenomenon, we consider comparing the class-level domain discrepancy taking source labels into account. Two local versions of $\mathcal{A}$-distance are consequently introduced as
\begin{equation}\label{eq:A_dist_class}
\small
\setlength\jot{0pt}
\setlength\abovedisplayskip{2pt plus 3pt minus 7pt}
\setlength\belowdisplayskip{2pt plus 3pt minus 7pt}
\begin{aligned}
&d_\mathcal{A}^{{\kern 1pt} w}=\frac{1}{C}\sum_i^Cd_\mathcal{A}(\mathcal{P}_s^i,\hat{\mathcal{P}}_s^i)\\
&d_\mathcal{A}^{{\kern 1pt} b}=\frac{1}{C(C-1)}\sum_{i=1}^C\sum_{j=1,j\neq i}^Cd_\mathcal{A}(\mathcal{P}_s^i,\mathcal{P}_s^j) {\kern 4pt}
\end{aligned}\\,
\end{equation}
where $d_\mathcal{A}^{{\kern 1pt} w}$ and $d_\mathcal{A}^{{\kern 1pt} b}$ quantifies the within- and between-class divergence, respectively. The superscript in $\mathcal{P}_s$ denotes a specific class in $C$ classes. In particular, considering the fact that, if data can be classified reasonably, it should have small within-class divergence and large between-class divergence. Therefore, $J_s=d_\mathcal{A}^{{\kern 1pt} w}/d_\mathcal{A}^{{\kern 1pt} b}$ is used to score the overall class-level domain discrepancy. Fig.~\ref{fig:A_dist}(c) shows the value of $J_s$ over the same DA tasks. At this time, the classification accuracy shows a similar trend with the $J_s$ measure. Our analysis justifies the tensor subspace well preserves the discriminative power of source domain.

\begin{figure*}[!tb]
	\centering
	\subfloat[NA]{\includegraphics[width=0.2\linewidth]{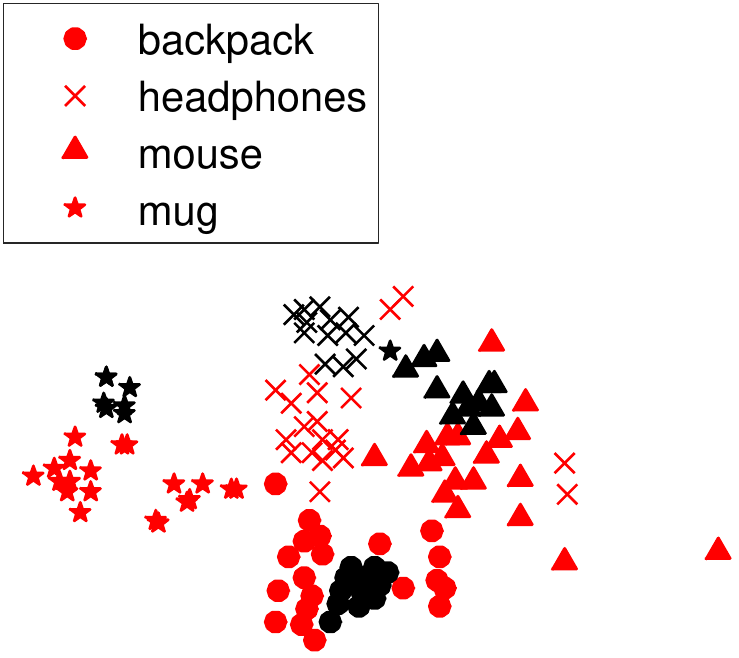}}\hspace{20pt}
	\subfloat[CORAL]{\includegraphics[width=0.16\linewidth]{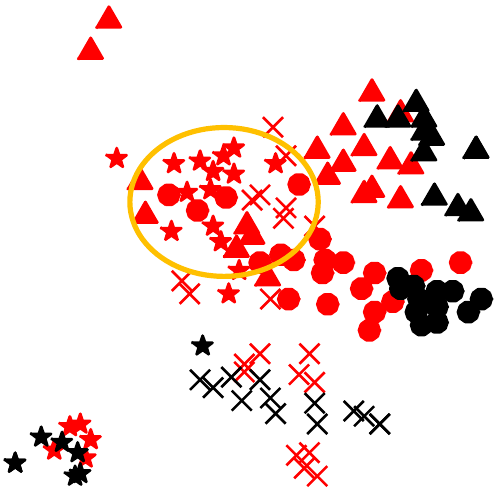}}\hspace{20pt}
	\subfloat[NTSL]{\includegraphics[width=0.2\linewidth]{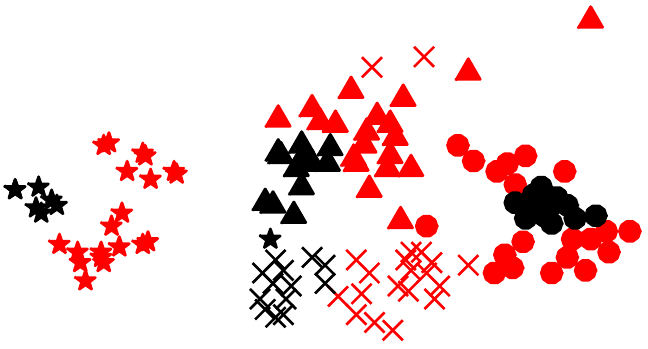}}\hspace{20pt}
	\subfloat[TAISL]{\includegraphics[width=0.18\linewidth]{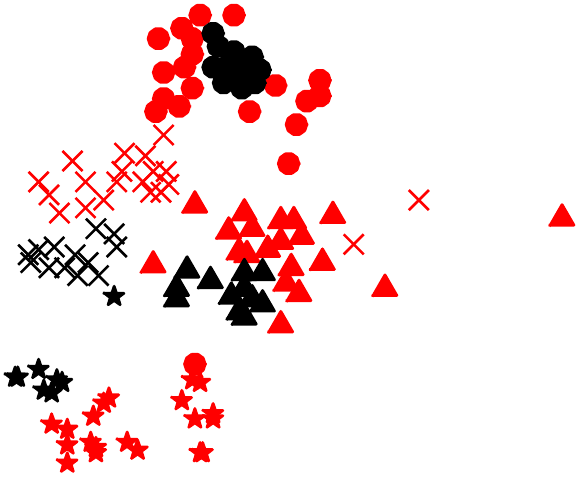}}
	\caption{Class-level data visualization using t-SNE~\cite{van2008visualizing} of different methods on the DA task of C (red) $\rightarrow$ D (black). 4 classes are chosen for better visualization. For CORAL, the data coming from the source domain tend to overlap with each other after the adaptation, a phenomenon we call \textit{over-adaptation}. (Best viewed in color.)}
	\label{fig:tsne}
\end{figure*}

To give a more intuitive illustration, the data distributions are visualized in Fig.~\ref{fig:tsne}. Indeed, the problem occurs during the transfer of source domain. As per the yellow circle in Fig.~\ref{fig:tsne}(b), different classes of the source data are overlapped after the adaptation. We call this phenomenon \textit{over-adaptation}. According to a recent study~\cite{wen2016discriminative}, there is a plausible explanation.~\cite{wen2016discriminative} shows that the feature distributions learned by CNNs are relatively ``fat''---the within-class variance is large, while the between-class margin is small. Hence, a slight disturbance would cause the overlaps among different classes. In CORAL, the disturbance perhaps boils down to the inexact estimation of covariance matrices caused by high feature dimensionality and limited source data. In contrast, as shown in Fig.~\ref{fig:tsne}(c)-(d), our approach naturally passes the discriminative power of source domain. Notice that, though the adaptation seems not perfect as target data are only aligned close to the source, the margins of different classes are clear so that there still has a high probability for target data to be classified correctly.

\begin{figure*}[!tb]
	\centering
	\subfloat[]{\includegraphics[width=0.349\linewidth]{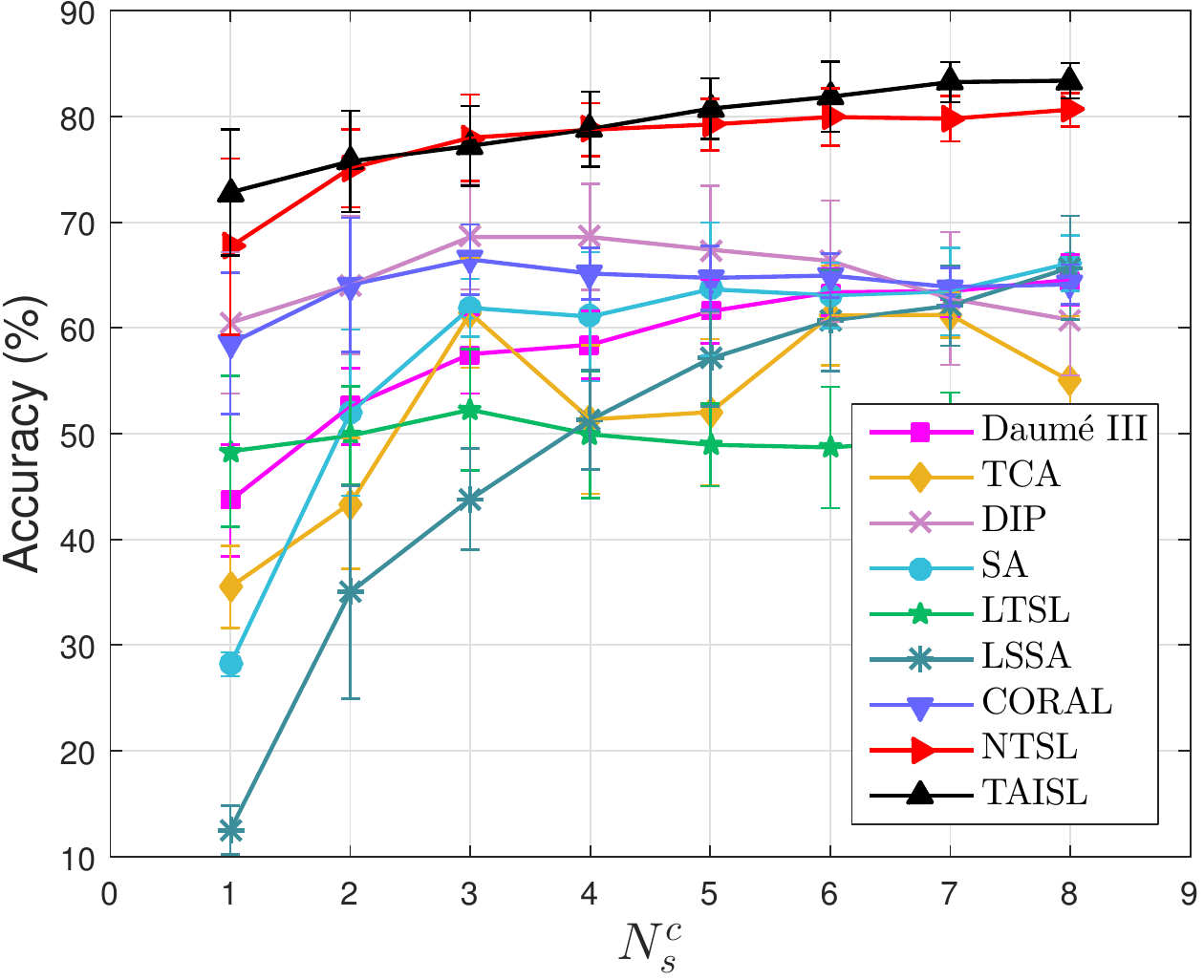}}
	\subfloat[]{\includegraphics[width=0.35\linewidth]{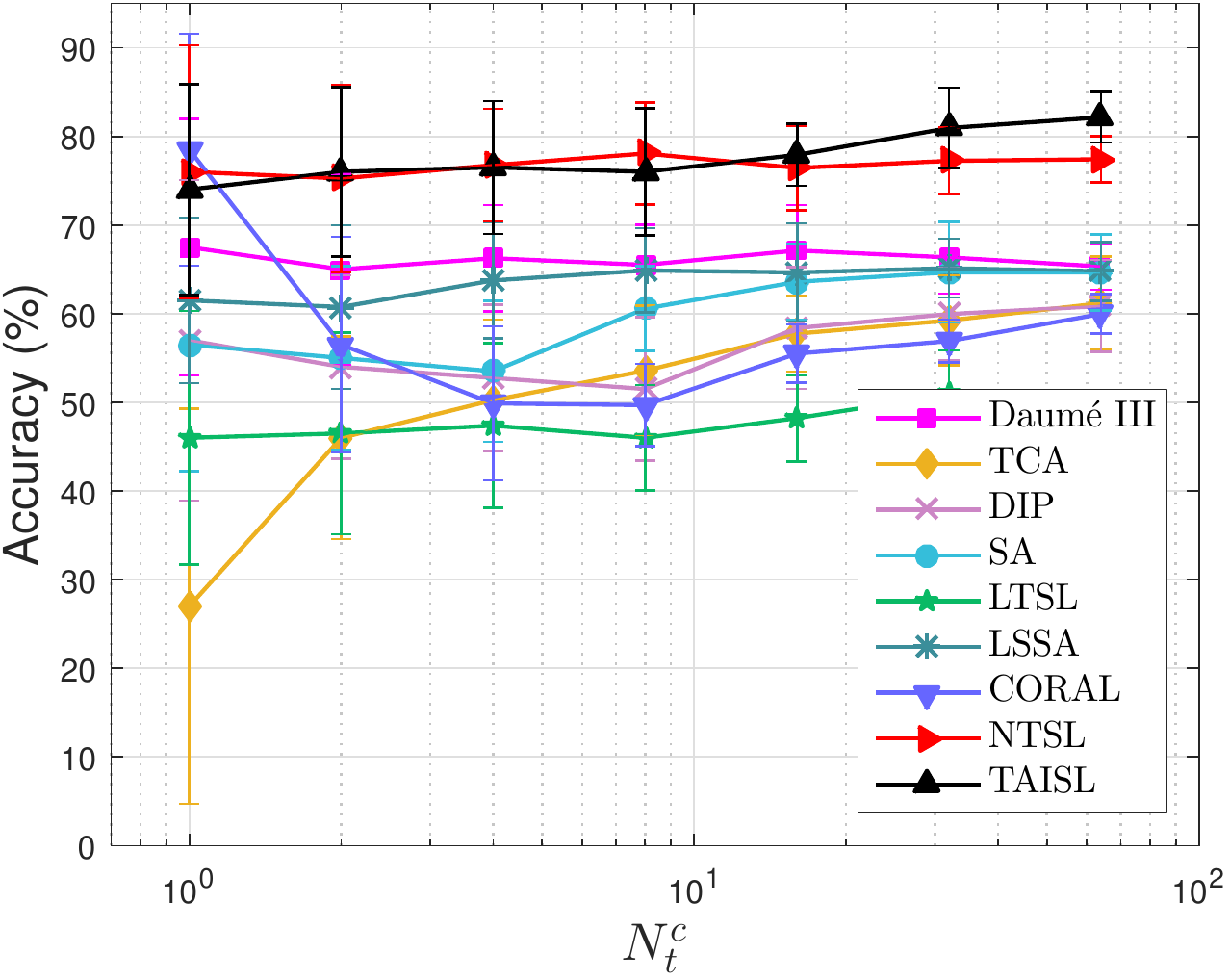}}\\
	\subfloat[]{\includegraphics[width=0.35\linewidth]{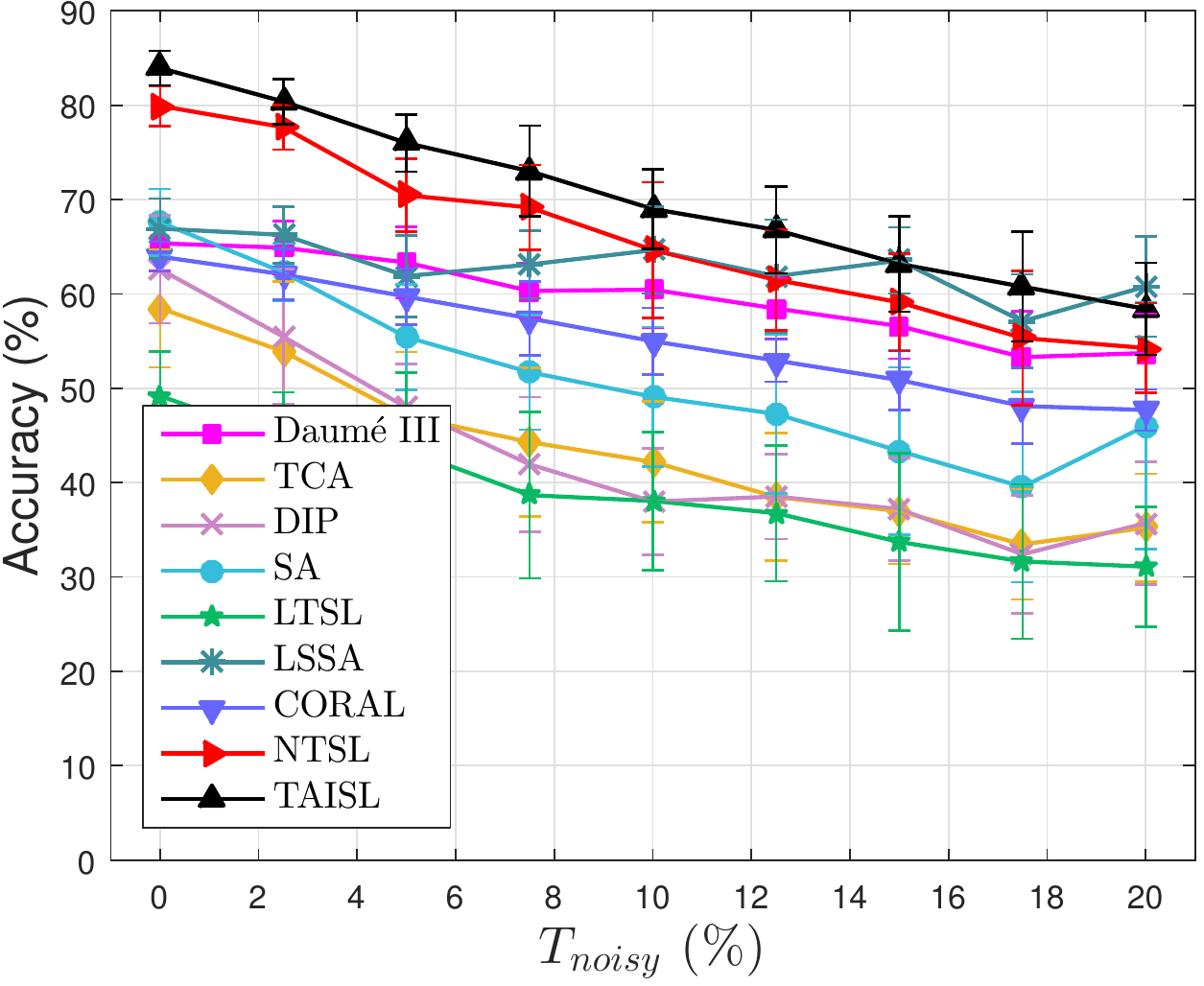}}
	\subfloat[]{\includegraphics[width=0.3495\linewidth]{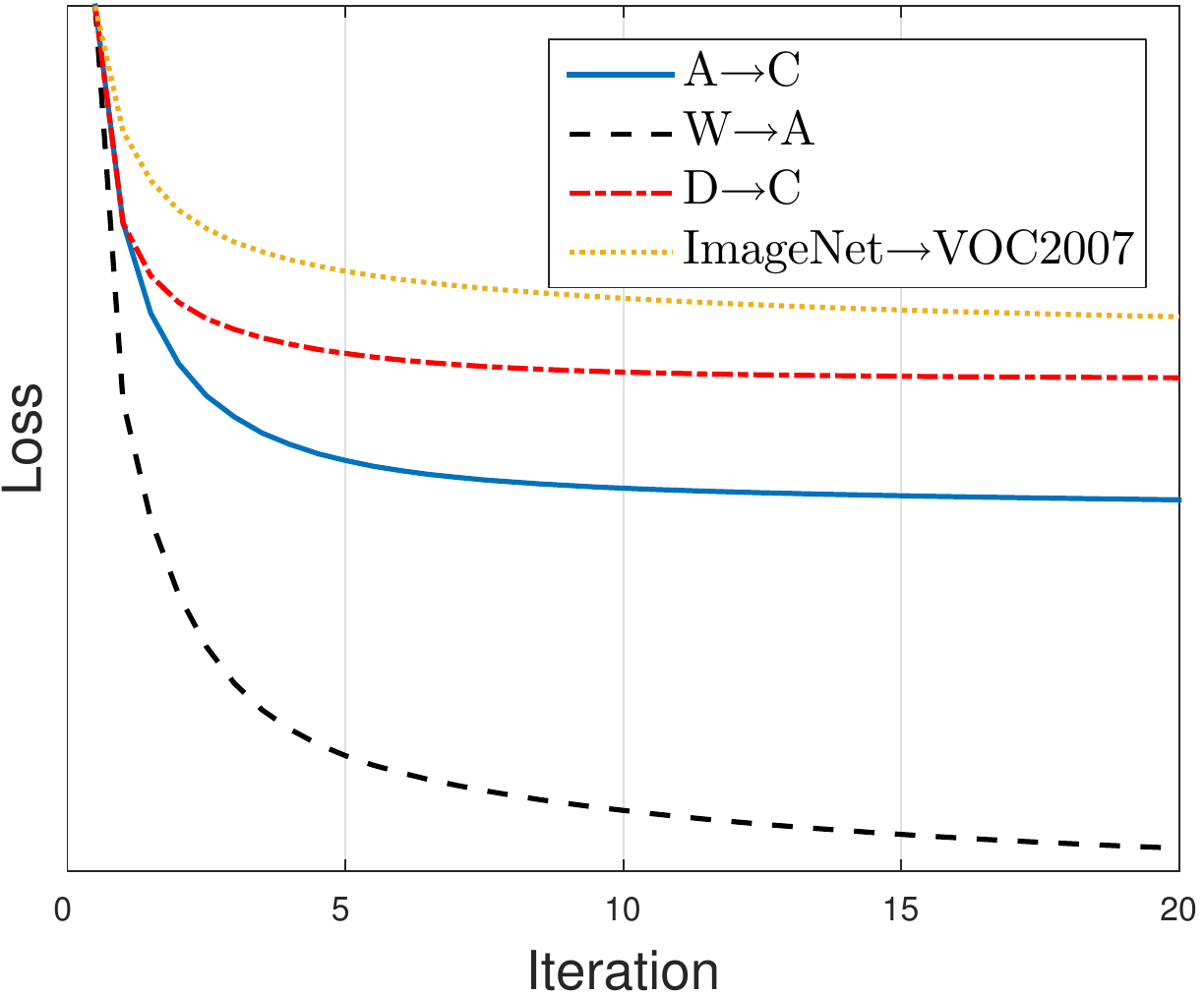}}
	\caption{Adaptation on D$\rightarrow$C with (a) varying number of source data per class $N_s^c$, (b) a varying number of target data per class $N_t^c$, and (c) noisy source labels. (d) Empirical convergence analysis of TAISL over several DA tasks.}
	\label{fig:acc_num_noisy_cvgs}
\end{figure*}

\paragraph{Adaptation with limited source/target data.} One of the important features of TAISL in practice is the small amount of training data required. In other words, one can characterize a domain, and thus adapt a pre-trained classifier, with very limited data. To demostrate this point, we evaluate the classification accuracy while varying the number of source/target data used for adaptation. The DA task of $D\rightarrow C$ is used. Concretely, we first fix the number of target data and, respectively, randomly choose from $1$ to $8$ source samples per category. In turn, we fix the number of source data to $8$ per category and set the target samples per category to $2^k, k = 0, 1, 2, ..., 7$. Fig.~\ref{fig:acc_num_noisy_cvgs}(a)-(b) illustrate the results of different approaches. It can be observed that, our approach demonstrates very stable classification performance, while other comparing methods is sensitive to the number of source samples used. Meanwhile, the number of target data seems not to have much impact on the classification accuracy, because in general one prefers to transfer the source domain so that the target domain does not change notably. It is worth noting that TAISL works even with only one source sample per category, which suggests that it can be applied for effective small-sample-size adaptation.

\paragraph{Adaptation with noisy labels.} Recent studies~\cite{Zhao15CP} demonstrate that tensor representations are inherently robust to noise. To further justify this in the context of DA, we randomly replace the source data with samples that have different labels. We gradually increase the percentage of noisy data $T_{noisy}$ from $0\%$ to $20\%$ and monitor the degradation of classification accuracy. As shown in Fig.~\ref{fig:acc_num_noisy_cvgs}(c), TAISL consistently demonstrates superior classification performance over other approaches.

\paragraph{Convergence analysis and efficiency comparison.} In this part, we empirically analyze the convergence behavior of TAISL. Fig.~\ref{fig:acc_num_noisy_cvgs}(d) shows the change of loss function as the iteration increases. It can be observed that the optimization generally converges in about $10$ iterations. In addition, we also compare the efficiency of different approaches. The average evaluation time of each trial is reported. According to Table~\ref{tab:time}, the efficiency of TAISL is competitive too. TCA and LSSA are fast, because these two methods adopt kernel tricks to avoid high-dimensional computation implicitly. In general, learning a tensor subspace is faster than a vector subspace in the high-dimensional case.

\paragraph{Recognition results.} Quantitative results are listed in Table~\ref{tab:offcal10_vgg_vd_16_res}. It shows that our approach is on par with or outperforms other related and state-of-the-art methods in terms of both average accuracy and standard deviations. Note that conventional methods that directly adapt vector-form convolutional activations sometimes have a negative effect on the classification, even falling behind the baseline NA. The main reason perhaps is the inexact estimation of a large amount of parameters. For instance, in many subspace-based approaches, one needs to estimate a flattened subspace from the covariance matrix. Given a data matrix $\mat{A}\in\mathbb{R}^{d\times n}$ with dimension $d$ and $n$ samples, its covariance matrix is estimated as $AA^T$. Notice that $rank(A)=rank(AA^T)=rank(A^TA)\leq \min(d, n-1)$. If $d\gg n$, the vector subspace will only be spanned by less than $n$ eigenvectors. In addition, one also suffers from the problem of biased estimation~\cite{wang2015beyond} (large eigenvalues turn larger, small ones turn smaller) when $d\gg n$. Hence, such vector subspaces are unreliable. In contrast, our approach avoids this problem due to the mode-wise parameters estimation.

\begin{table}[!t] \footnotesize
	\centering
	\addtolength{\tabcolsep}{1.2pt}
	\renewcommand\arraystretch{1.05}
	\begin{tabular}{l|cccccccccc}
		\hline
		Method & Daum\'{e} III & TCA & GFK & DIP & SA \\
		Time   & \small{0.06} & \small{0.05} & \small{3.94} & \small{9.09} & \small{3.40} \\
		\hline
		Method & LTSL & LSSA & CORAL & NTSL & TAISL \\
		Time   & \small{12.34} & \small{0.59} & \small{14.81} & \small{0.16} & \small{0.92} \\
		\hline
	\end{tabular}
	\vspace{-5pt}
	\caption{Average evaluation time (min) of each trial of different methods on A$\rightarrow$C. (Matlab 2016a, OS: OS X 64-bit, CPU: Intel i5 2.9GHz, RAM: 8 GB)}
	\label{tab:time}
\end{table}

\begin{figure}[!t]
	\vspace{-5pt}
	\centering
	\subfloat[A$\rightarrow$C]{\includegraphics[width=0.48\linewidth]{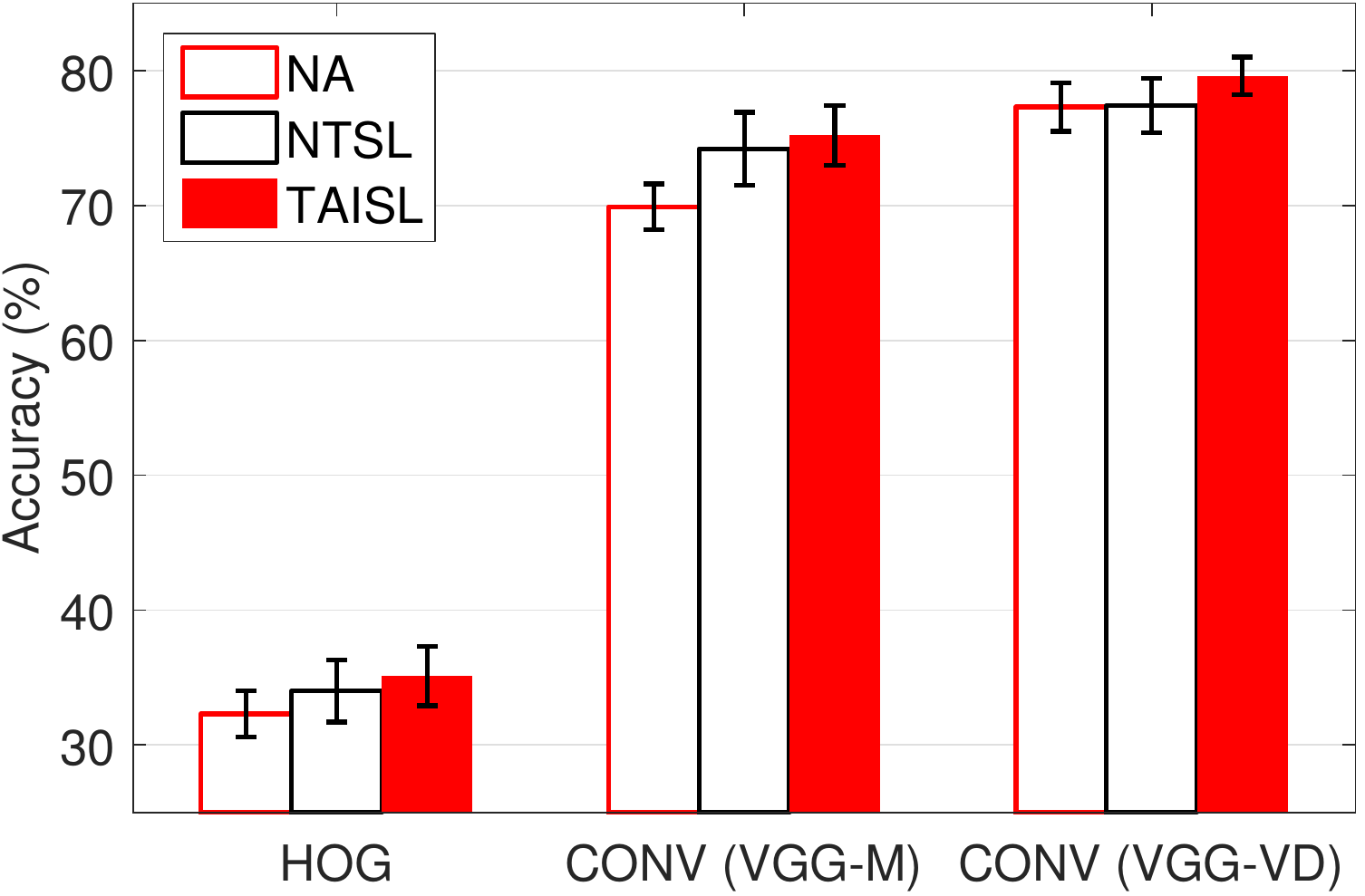}} \kern 5pt
	\subfloat[W$\rightarrow$A]{\includegraphics[width=0.48\linewidth]{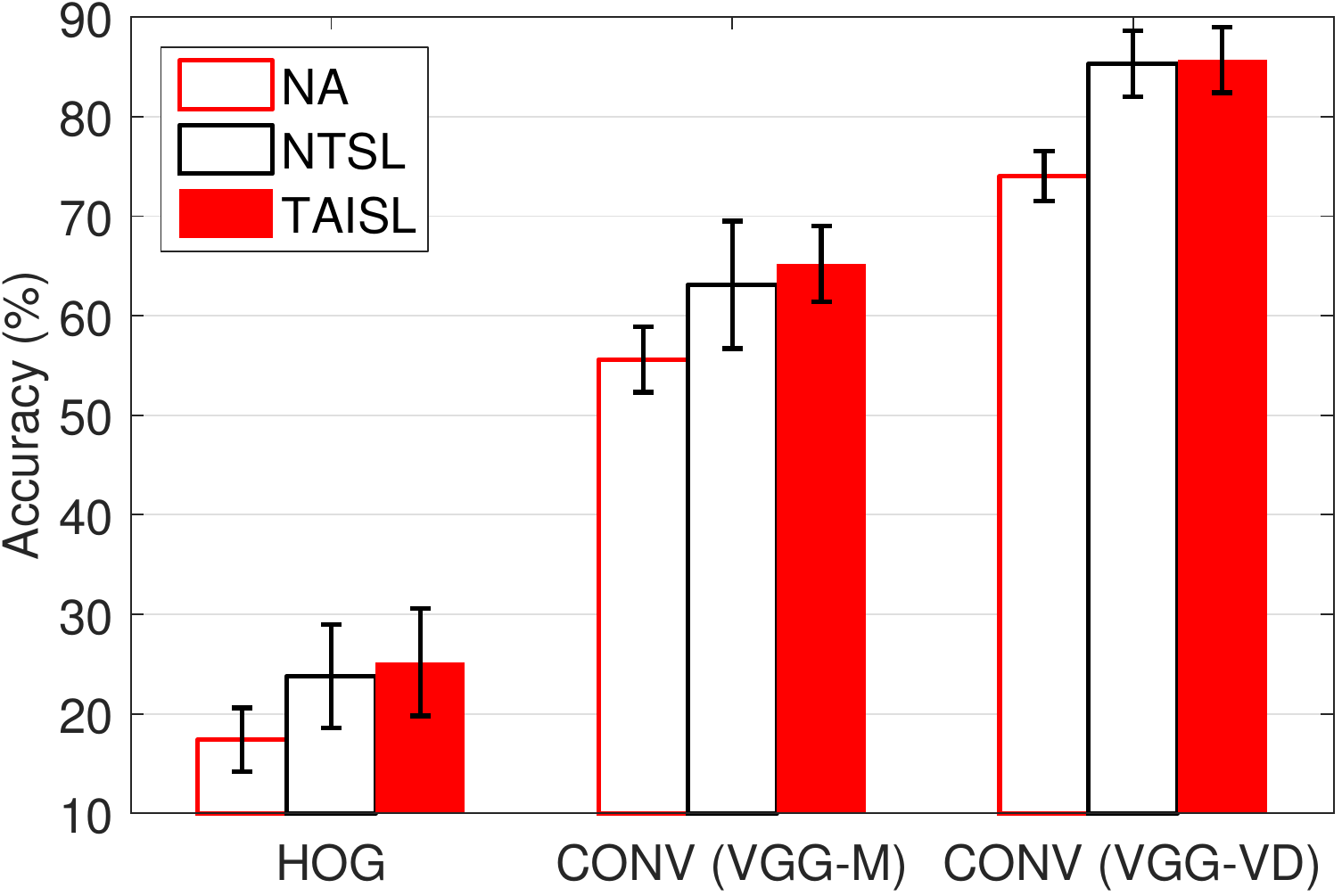}}
	\caption{Adaptation accuracy of three types of tensor representations on two DA tasks.}
	\label{fig:feat}
\end{figure}

\begin{table*}[!tb] \footnotesize
	\centering
	\addtolength{\tabcolsep}{-1.1pt}
	\renewcommand\arraystretch{1.1}
	\begin{tabular}{l|ccccccccccccc}
		\hline
		Method & \scriptsize A$\rightarrow$C  & \scriptsize C$\rightarrow$A  & \scriptsize A$\rightarrow$D  & \scriptsize D$\rightarrow$A  & \scriptsize A$\rightarrow$W  & \scriptsize W$\rightarrow$A  & \scriptsize C$\rightarrow$D  & \scriptsize D$\rightarrow$C  & \scriptsize C$\rightarrow$W  & \scriptsize W$\rightarrow$C  & \scriptsize D$\rightarrow$W  & \scriptsize  W$\rightarrow$D & \textsc{mean}\\
		\hline
		NA & 77.3\sz({1.8}) & 89.0\sz({2.0}) & 82.8\sz({2.2}) & 81.1\sz({1.9}) & 74.6\sz({3.1}) & 74.0\sz({2.5}) & 86.2\sz({4.0}) & 70.5\sz({1.9}) & 79.4\sz({2.7}) & 63.7\sz({2.1}) & 91.1\sz({1.7}) & 94.9\sz({2.4}) & 80.4\\
		PCA & 36.7\sz({3.0}) & 57.7\sz({5.2}) & 23.5\sz({8.1}) & 50.2\sz({4.8}) & 18.6\sz({6.2}) & 51.2\sz({6.0}) & 29.9\sz({7.8}) & 51.0\sz({3.3}) & 26.4\sz({6.7}) & 51.6\sz({3.6}) & 49.5\sz({4.2}) & 50.8\sz({7.0}) & 41.4\\
		Daum\'{e} III & 73.1\sz({1.5}) & 85.9\sz({2.5}) & 70.9\sz({3.7}) & 59.9\sz({7.1}) & 70.6\sz({3.5}) & 68.7\sz({4.4}) & 81.4\sz({4.1}) & 56.2\sz({6.4}) & 75.4\sz({4.0}) & 59.6\sz({2.6}) & 81.5\sz({2.7}) & 86.5\sz({4.9}) & 72.5\\
		TCA & 56.7\sz({4.5}) & 78.1\sz({6.1}) & 59.9\sz({6.7}) & 61.2\sz({4.2}) & 55.5\sz({6.4}) & 68.3\sz({4.1}) & 74.3\sz({5.2}) & 51.9\sz({2.2}) & 69.0\sz({6.6}) & 54.7\sz({3.8}) & 89.8\sz({2.2}) & 90.6\sz({3.2}) & 67.5\\
		GFK & 75.1\sz({3.9}) & 87.6\sz({2.3}) & 81.4\sz({4.3}) & \red{90.4\sz({1.4})} & 74.3\sz({5.2}) & 84.0\sz({4.4}) & 84.8\sz({4.5}) & \red{82.2\sz({2.4})} & \red{81.9\sz({4.9})} & 79.1\sz({2.7}) & 92.8\sz({2.2}) & 95.2\sz({2.2}) & 84.1\\
		DIP & 59.8\sz({5.7}) & 84.8\sz({4.3}) & 52.2\sz({8.1}) & 76.4\sz({3.7}) & 45.5\sz({9.1}) & 69.3\sz({6.9}) & 82.8\sz({7.7}) & 61.9\sz({6.3}) & 73.5\sz({4.9}) & 65.2\sz({4.5}) & 90.9\sz({2.3}) & 94.1\sz({3.1}) & 71.4\\
		SA & 67.7\sz({4.2}) & 82.0\sz({2.6}) & 67.8\sz({4.8}) & 77.4\sz({6.0}) & 61.1\sz({5.1}) & 80.1\sz({4.3}) & 73.7\sz({4.3}) & 66.9\sz({3.3}) & 65.9\sz({4.0}) & 70.4\sz({4.1}) & 87.3\sz({3.1}) & 91.1\sz({3.3}) & 74.3\\
		LTSL & 70.2\sz({2.4}) & 87.5\sz({2.8}) & 77.7\sz({4.6}) & 69.2\sz({4.5}) & 66.7\sz({4.6}) & 66.6\sz({5.7}) & 82.3\sz({4.1}) & 60.8\sz({3.1}) & 75.3\sz({4.2}) & 59.1\sz({4.4}) & 86.0\sz({2.9}) & 90.0\sz({3.8}) & 74.3\\
		LSSA & \textbf{80.3\sz({2.3})} & 86.4\sz({1.7}) & \textbf{90.9\sz({1.7})} & \textbf{92.3\sz({0.6})} & \textbf{84.0\sz({1.7})} & \textbf{86.6\sz({4.5})} & 73.5\sz({2.3}) & 65.9\sz({6.5}) & 45.4\sz({6.6}) & 29.5\sz({7.0}) & 93.4\sz({2.2}) & 85.8\sz({4.7}) & 76.2\\
		CORAL & 77.6\sz({1.2}) & 80.3\sz({1.9}) & 64.3\sz({2.9}) & 74.2\sz({2.2}) & 61.2\sz({2.4}) & 69.1\sz({2.6}) & 62.1\sz({3.0}) & 72.0\sz({1.7}) & 63.8\sz({3.1}) & 66.6\sz({2.2}) & 89.6\sz({1.6}) & 82.8\sz({2.8}) & 72.0\\
		\hline
		NTSL & 78.5\sz({2.3}) & \red{89.6\sz({2.2})} & 83.1\sz({3.3}) & 87.8\sz({1.4}) & 77.3\sz({3.1}) & \red{85.8\sz({2.8})} & \red{87.7\sz({2.9})} & 79.8\sz({1.5}) & 80.4\sz({3.8}) & \red{80.0\sz({2.0})} & \red{95.4\sz({1.4})} & \textbf{97.8\sz({1.7})} & \red{85.3}\\
		TAISL & \red{80.1\sz({1.4})} & \textbf{90.0\sz({1.9})} & \red{85.1\sz({2.2})} & 87.6\sz({2.1}) & \red{77.9\sz({2.6})} & 85.6\sz({3.5}) & \textbf{90.6\sz({1.9})} & \textbf{84.0\sz({1.0})} & \textbf{85.3\sz({3.1})} & \textbf{82.6\sz({2.2})} & \textbf{95.9\sz({1.0})} & \red{97.7\sz({1.5})} & \textbf{86.9}\\
		\hline
	\end{tabular}
	\caption{Average multi-class recognition accuracy (\%) on Office--Caltech10 dataset over 20 trials. The highest accuracy in each column is boldfaced, the second best is marked in red, and standard deviations are shown in parentheses.}
	\label{tab:offcal10_vgg_vd_16_res}
\end{table*}

\begin{table*}[!tb] \footnotesize
	\centering
	\addtolength{\tabcolsep}{1.15pt}
	\renewcommand\arraystretch{1.1}
	\begin{tabular}{l|cccccccccccc}
		\hline
		Method & \footnotesize aero  & \footnotesize bird  & \footnotesize bottle  & \footnotesize cat  & \footnotesize cow & \footnotesize table  & \footnotesize mbike  & \footnotesize person & \footnotesize sheep  & \footnotesize tv & mAP\\
		\hline
		NA    & 66.4\sz({2.1})  & 65.6\sz({4.0})  & 29.5\sz({2.1})  & 70.6\sz({3.4})  & 30.3\sz({8.0}) & 35.7\sz({5.5}) & 47.0\sz({8.0}) & 69.3\sz({2.9}) & 44.9\sz({6.9}) & 56.4\sz({3.3}) & 51.6\\
		PCA   & 28.9\sz({5.8})  & 30.2\sz({3.9})  & 23.3\sz({5.2})  & 44.9\sz({4.5})  & 6.0\sz({1.8}) & 29.0\sz({6.9}) & 25.0\sz({5.0}) & 70.2\sz({1.9}) & 11.7\sz({3.5}) & 29.0\sz({6.6}) & 29.8\\
		Daum\'{e} III & 64.1\sz({3.7})  & 59.7\sz({7.4})  & 26.6\sz({3.3})  & 65.7\sz({5.3})  & 26.9\sz({8.5}) & 30.0\sz({5.4}) & 40.5\sz({6.6}) & 68.5\sz({2.5}) & 37.7\sz({7.4}) & 51.9\sz({4.4}) & 47.2\\
		TCA   & 43.2\sz({9.8})  & 44.4\sz({10.5}) & 20.7\sz({1.7})  & 56.7\sz({8.2})  & 16.9\sz({6.3}) & 27.6\sz({8.7}) & 31.8\sz({10.2}) & 58.1\sz({5.7}) & 22.7\sz({8.0}) & 33.6\sz({10.2}) & 35.6\\
		GFK   & 70.0\sz({6.9})  & 74.6\sz({3.8})  & 32.5\sz({4.4})  & 73.1\sz({6.5})  & 28.9\sz({5.3}) & 48.3\sz({10.4}) & 58.3\sz({4.8}) & \textbf{75.8\sz({3.6})} & 52.5\sz({4.8}) & 57.1\sz({4.5}) & 57.1\\
		DIP   & 69.8\sz({5.5})  & \red{78.4\sz({4.6})}  & 29.1\sz({5.0})  & \red{75.9\sz({3.7})}  & 25.5\sz({5.0}) & 42.2\sz({8.1}) & 56.3\sz({5.7}) & \red{73.5\sz({3.1})} & 48.9\sz({4.3}) & 59.4\sz({5.2}) & 55.9\\
		SA    & 64.4\sz({10.1})  & 69.3\sz({5.4})  & 34.4\sz({4.6})  & 67.4\sz({4.9})  & 18.4\sz({6.6}) & 36.9\sz({12.8}) & 53.7\sz({10.9}) & 68.9\sz({2.4}) & 31.4\sz({10.2}) & 55.2\sz({5.7}) & 50.0\\
		LTSL  & 56.9\sz({10.4})  & 61.0\sz({7.7})  & 34.9\sz({6.2})  & 70.8\sz({8.8})  & 21.9\sz({6.3}) & 43.7\sz({12.4}) & 52.5\sz({10.7}) & 69.9\sz({4.3}) & 38.2\sz({9.5}) & 54.0\sz({7.5}) & 50.4\\
		LSSA  & \textbf{78.7\sz({2.0})}  & \textbf{79.7\sz({1.2})} & \textbf{38.4\sz({4.6})} & \textbf{81.7\sz({0.5})} & 29.5\sz({1.9}) & 33.7\sz({3.4}) & 56.3\sz({9.3}) & 51.2\sz({2.0}) & 32.5\sz({10.6}) & 51.6\sz({4.6}) & 53.3\\
		CORAL & 71.4\sz({3.3})  & 71.7\sz({3.6})  & 35.2\sz({2.4})  & 72.0\sz({4.3})  & \textbf{36.0\sz({5.7})} & 40.6\sz({6.7}) & 57.3\sz({5.6}) & 67.6\sz({2.0}) & \textbf{54.8\sz({2.9})} & 56.9\sz({3.6}) & 56.6\\
		\hline
		NTSL  & 76.3\sz({4.3})  & 71.0\sz({3.9})  & 35.7\sz({3.7})  & 71.3\sz({3.2})  & \red{34.7\sz({9.8})} & \red{49.8\sz({10.4})} & \red{59.7\sz({10.2})} & 72.0\sz({4.6})  & 53.4\sz({6.0}) & \red{60.2\sz({3.5})} & \red{58.4}\\
		TAISL & \red{76.4\sz({5.1})}  & 71.6\sz({3.1})  & \red{36.7\sz({3.5})}  & 72.0\sz({2.1})  & 33.3\sz({6.6}) & \textbf{50.7\sz({10.0})} & \textbf{60.3\sz({8.7})} & 72.2\sz({3.8}) & \red{53.6\sz({5.6})} & \textbf{60.4\sz({3.5})} & \textbf{58.7}\\
		\hline
	\end{tabular}
	\caption{Average precision (\%) on ImageNet--VOC2007 dataset over 10 trials. The highest performance in each column is boldfaced, the second best is marked in red, and standard deviations are shown in parentheses.}
	\label{tab:imagenet_voc_vgg_vd_16_res}
\end{table*}

\subsection{Evaluation on the ImageNet--VOC2007 dataset}
\label{subsec:eval_iv}

Here we evaluate our approach under a more challenging dataset than OC10. As aforementioned, VOC2007 is a multi-label dataset, so many images contain multiple labels. Results are listed in Table~\ref{tab:imagenet_voc_vgg_vd_16_res}. Due to the space limitation, we show only results of 10 categories (additional results are attached in the Supplementary). We observe that TAISL still demonstrates the best overall classification performance among comparing approaches. We also notice that NTSL and TAISL show comparable results. We conjecture that,
since the target domain contains too many noisy labels, it will be hard to determine a global alignment that just matches class-level differences. As a result, the alignment may not work the way it should. In addition, according to Tables~\ref{tab:offcal10_vgg_vd_16_res} and~\ref{tab:imagenet_voc_vgg_vd_16_res}, LSSA shows superior accuracy than ours over several DA tasks/categories. It makes sense because LSSA works at different levels with further non-linearity and samples reweighting. However, non-linearity is a double-edged sword. It can improve the accuracy in some situations, while sometimes it may not. For instance, the accuracy of LSSA drops significantly on the W$\rightarrow$C task.

\subsection{Evaluation with other tensor representations}
\label{subsec:eval_feat}

Finally, we evaluate other types of tensor representations to validate the generality of our approach. We do not limit the representation from deep learning features. Other shallow tensor features also can be adapted by our approach.  Specifically, the improved HOG feature~\cite{felzenszwalb2010dpm} and convolutional activations extracted from the CONV5 layer of VGG--M~\cite{Chatfield14} model are further utilized and evaluated on two DA tasks from the OC10 dataset. Results are shown in Fig.~\ref{fig:feat}. We notice that TAISL consistently improves the recognition accuracy with various tensor representations. In addition, a tendency shows that, the better feature representations are, the higher the baseline achieves, which implies a fundamental rule of domain-invariant feature representations for DA.

\section{Conclusion}
\label{sec:conclu}

Practical application of machine learning techniques often gives rise to situations where domain adaptation is required, either because acquiring the perfect training data is difficult, the domain shift is unpredictable, or simply because it is easier to re-use an existing model than to train a new one. This is particularly true for CNNs as the training time and data requirements are significant.

The DA method proposed in this work is applicable in the case where a tensor representation naturally captures information that would be difficult to represent using vector arithmetic, but also benefits from the fact that it uses a lower-dimensional representation to achieve DA, and thus is less susceptible to noise.  We have shown experimentally that it outperforms the state of the art, most interestingly for CNN DA, but is also much more efficient.

In  future work, discriminative information from source data may be employed for learning a more powerful invariant tensor subspace.

\newpage
\onecolumn

\section*{Appendix}

In this Appendix, we provide more details that are not included in the main text due to the page limitation. In particular, we supplement the following content on
\squishlist
\item how to implement the optimization of our approach efficiently;
\item how to perform spatial pooling normalization to convolutional activations; we only briefly mention this procedure in Section~5.1 of the main text;
\item detailed introduction regarding used datasets;
\item additional results evaluated on Office and ImageNet--VOC2007 datasets; 
\item parameters sensitivity.
\squishend

\section{Towards efficient optimization}

In this section, we will reveal several important details towards efficient practical implementations. Note that $\mathcal{X}_s\in{\mathbb{R}^{n_1 \times ... \times n_K \times N_s}}$ is a ($K+1$)-mode tensor, the unfolding matrix $\mat{X}_{s(k)}$ is of size $n_k \times n_{\setminus k}N_s$, where $n_{\setminus k} = n_1\cdots n_{k-1} n_{k+1}\cdots n_K$. When computing $\mat{Q}^{(k)} = \mat{X}_{s(k)}\mat{M}^T_{\setminus k}$ in Eq.~(13), $\mat{M}^T_{\setminus k}$ will be of size $n_{\setminus k}N_s \times n_{\setminus k}N_s$, which is extremely large and consume a huge amount of memory to store. In fact, such a matrix even cannot be constructed in a general-purpose computer. To alleviate this, we choose to solve an equivalent optimization problem by reformulating Eq.~(13) into its sum form as
\begin{equation}\label{eq:tasl_sum}
\begin{aligned}
\min_{{\mat{\scriptstyle M}^{(k)}}} & {\kern 4pt} \sum_{n=1}^{N_s} \|\mat{M}^{(k)}\mat{Q}^n_{(k)} - \mat{Y}^n_{(k)}\|^2_F - \lambda \|\mat{M}^{(k)}\mat{X}^n_{s(k)}\|_F^2 {\kern 4pt}\\
{\rm{s.t.}} & {\kern 4pt} \forall k, {\kern 2pt} {\mat{M}^{(k)}}{\mat{M}^{(k)T}} = \mat{I}
\end{aligned}\\,
\end{equation}
where
\begin{equation}\label{eq:tasl_sum1}
\begin{aligned}
\mat{Q}^n_{(k)} & = \mat{X}^n_{s(k)}\hat{\mat{M}}^T_{\setminus k} {\kern 4pt}\\
\hat{\mat{M}}^T_{\setminus k} & =\mat{M}^{(K)}\otimes\cdots\otimes\mat{M}^{(k+1)}\otimes\mat{M}^{(k-1)}\otimes\cdots\otimes\mat{M}^{(1)} {\kern 4pt}\\
\end{aligned}\\,
\end{equation}
$\mat{Y}^n_{(k)}=\mat{Y}_{(k)}(:, :, n)$ ($\mat{Y}_{(k)}$ has been reshaped into the size of $n_k\times n_{\setminus k}\times N_s$), and $\mat{X}^n_{s(k)}$ denotes the $k$-th mode unfolding matrix of $\mathcal{X}_s^n$. In following expressions, we denote $\mat{Q}^n_{(k)}$, $\mat{Y}^n_{(k)}$, and $\mat{X}^n_{s(k)}$ by $\mat{Q}_n$, $\mat{Y}_n$, and $\mat{X}_n$ for short, respectively. By replacing $\mat{M}^{(k)T}$ with $\mat{P}$, we arrive at
\begin{equation}\label{eq:tasl_sum2}
\begin{aligned}
\min_{{\mat{\scriptstyle P}}} & {\kern 4pt} \sum_{n=1}^{N_s} \|\mat{Q}_n^T\mat{P} - \mat{Y}_n^T\|^2_F - \lambda \|\mat{X}_n^T\mat{P}\|_F^2 {\kern 4pt}\\
{\rm{s.t.}} & {\kern 4pt} \forall k, {\kern 2pt} {\mat{P}^T}{\mat{P}} = \mat{I}
\end{aligned}\\.
\end{equation}
Considering that a standard solver needs the loss function $\mathcal{F}$ and its gradient $\partial\mathcal{F}/\partial\mat{P}$ as the input, we can compute them in the following way to speed up the optimization process. For the loss function $\mathcal{F}$, we have
\begin{equation}\label{eq:tasl_sum3}
\begin{aligned}
\mathcal{F} {\kern 4pt}
& = {\kern 4pt} \sum_{n=1}^{N_s} \|\mat{Q}_n^T\mat{P} - \mat{Y}_n^T\|^2_F - \lambda \|\mat{X}_n^T\mat{P}\|_F^2 {\kern 4pt}\\
& = {\kern 4pt} \sum_{n=1}^{N_s} Tr\left[(\mat{Q}_n^T\mat{P} - \mat{Y}_n^T)^T(\mat{Q}_n^T\mat{P} - \mat{Y}_n^T)\right] - \lambda \sum_{n=1}^{N_s} Tr\left[(\mat{X}_n^T\mat{P})^T(\mat{X}_n^T\mat{P})\right] {\kern 4pt}\\
& = {\kern 4pt} \sum_{n=1}^{N_s} Tr\left[\mat{P}^T\mat{Q}_n\mat{Q}_n^T\mat{P} - 2\mat{P}^T\mat{Q}_n\mat{Y}_n^T + \mat{Y}_n\mat{Y}_n^T\right] - \lambda \sum_{n=1}^{N_s} Tr\left[\mat{P}^T\mat{X}_n\mat{X}_n^T\mat{P}\right] {\kern 4pt}\\
& = {\kern 4pt} Tr\left[\mat{P}^T(\sum_{n=1}^{N_s}\mat{Q}_n\mat{Q}_n^T)\mat{P}\right] - 2Tr\left[\mat{P}^T(\sum_{n=1}^{N_s}\mat{Q}_n\mat{Y}_n^T)\right] + Tr\left[\sum_{n=1}^{N_s}\mat{Y}_n\mat{Y}_n^T\right] - \lambda Tr\left[\mat{P}^T(\sum_{n=1}^{N_s}\mat{X}_n\mat{X}_n^T)\mat{P}\right] {\kern 4pt}\\
\end{aligned}\\,
\end{equation}
where $Tr[\cdot]$ denotes the trace of matrix. For the gradient $\partial\mathcal{F}/\partial\mat{P}$, we have
\begin{equation}\label{eq:tasl_sum4}
\begin{aligned}
\partial\mathcal{F}/\partial\mat{P} {\kern 4pt}
& = {\kern 4pt} 2\sum_{n=1}^{N_s} \mat{Q}_n(\mat{Q}_n^T\mat{P} - \mat{Y}_n^T) - 2\lambda \sum_{n=1}^{N_s} \mat{X}_n\mat{X}_n^T\mat{P} {\kern 4pt}\\
& = {\kern 4pt} 2 (\sum_{n=1}^{N_s}\mat{Q}_n\mat{Q}_n^T)\mat{P} - 2\sum_{n=1}^{N_s}\mat{Q}_n\mat{Y}_n^T - 2\lambda (\sum_{n=1}^{N_s} \mat{X}_n\mat{X}_n^T)\mat{P} {\kern 4pt}\\
\end{aligned}\\.
\end{equation}
Notice that both $\mathcal{F}$ and $\partial\mathcal{F}/\partial\mat{P}$ share some components. As a consequence, we can precompute $\sum_{n=1}^{N_s}\mat{Q}_n\mat{Q}_n^T$, $\sum_{n=1}^{N_s}\mat{Q}_n\mat{Y}_n^T$, $\sum_{n=1}^{N_s}\mat{Y}_n\mat{Y}_n^T$, and $\sum_{n=1}^{N_s}\mat{X}_n\mat{X}_n^T$ before the $\mathcal{M}$-step optimization instead of directly feeding the original variables and iteratively looping over $N_s$ samples inside the optimization. Such a kind of precomputation speeds up the optimization significantly.

\section{Feature normalization with spatial pooling}

\begin{figure}[!b]
	\centering
	\includegraphics[width=8cm]{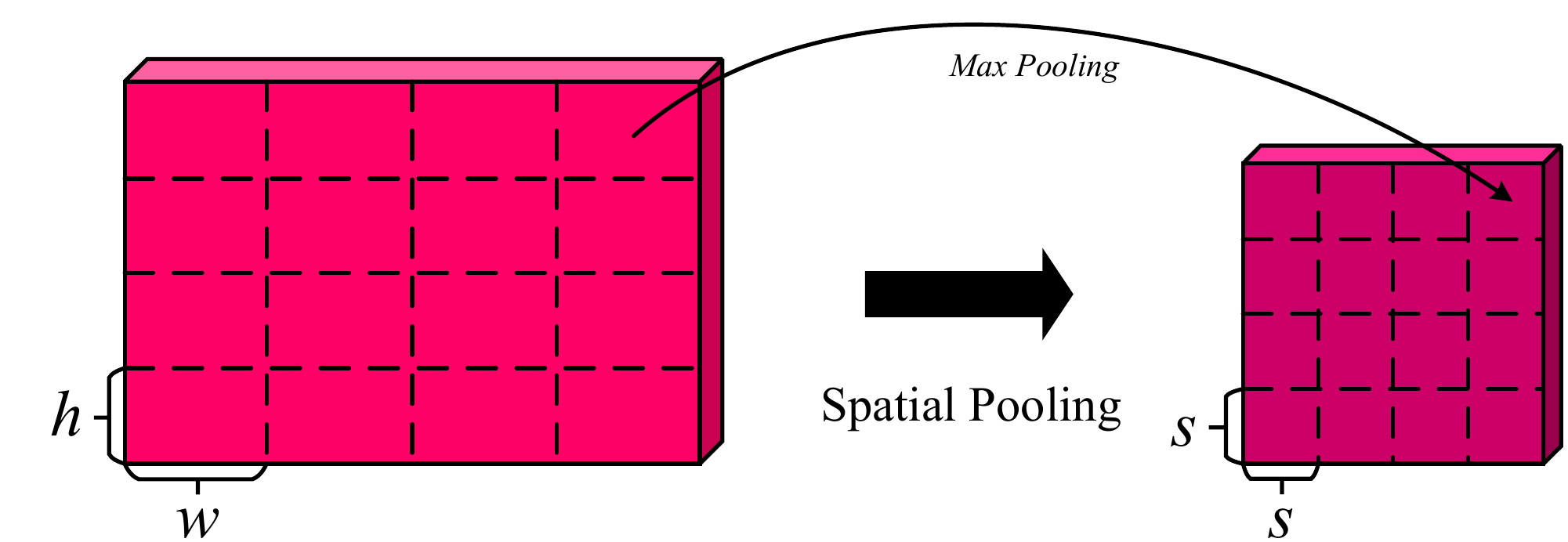}
	\caption{Illustration of spatial pooling normalization. Any size of convolutional representations will be normalized to a fixed-size tensor.}
	\label{fig:sp}
\end{figure}

Since we allow the input image to be of arbitrary size, a normalization step need to perform to ensure the consistency of dimensionality. The idea of spatial pooling is similar to the spatial pyramid pooling in~\cite{He2015}. The difference is that we do not pool pyramidally and do not vectorize the pooled activations, in order to preserve the spatial information. Intuitively, Fig.~\ref{fig:sp} illustrates this process. More concretely, convolutional activations are first equally divided into $N_{bin}$ bins along the spatial modes ($N_{bin}=16$ in Fig.~\ref{fig:sp}). Next, each bin with size of $h\times w$ is normalized to a $s\times s$ bin by max pooling. In our experiments, we set $N_{bin}=36$ and $s=1$.

\section{Datasets and protocol details}

\begin{figure}[!t]
	\centering
	\includegraphics[width=16cm]{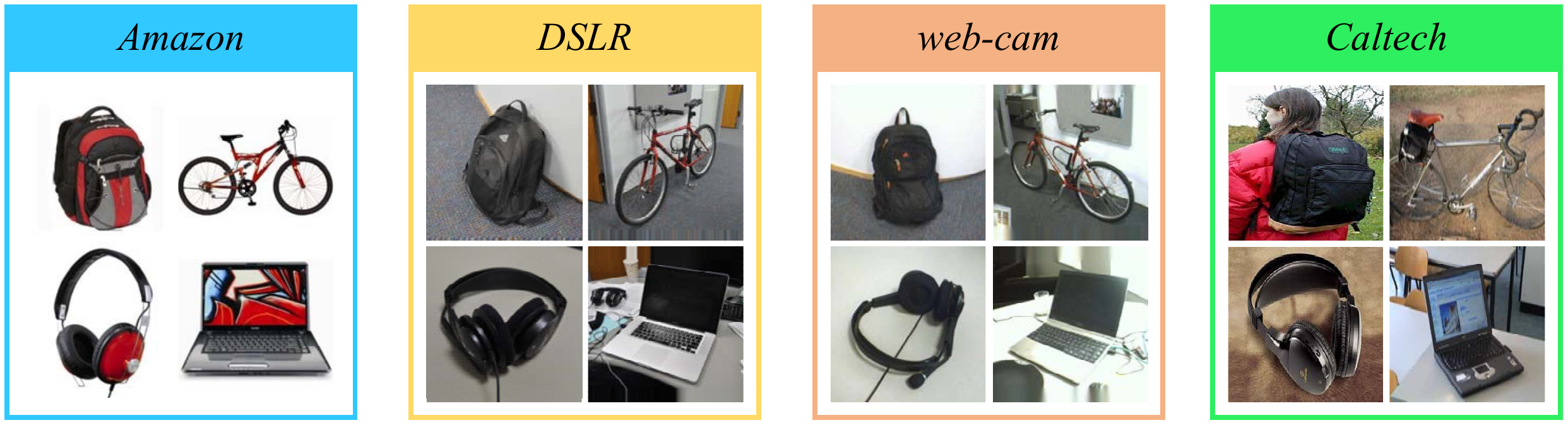}
	\caption{Some images from Office--Caltech10 dataset. 4 categories of \texttt{backpack}, \texttt{bike}, \texttt{headphone}, and \texttt{laptop computer} are selected.}
	\label{fig:offcal10}
\end{figure}

\paragraph{Office--Caltech10 dataset.} As mentioned in the main text,~\cite{gong12geodesic} extends Office~\cite{saenko2010adapting} dataset by adding another \textit{Caltech} domain. They select 10 common categories from four domains, including \textit{Amazon}, \textit{DSLR}, \textit{web-cam}, and \textit{Caltech}. \textit{Amazon} consists of images used in the online market, which shows the objects from a canonical viewpoint. \textit{DSLR} contains images captured with a high-resolution digital camera. Images in \textit{web-cam} are recorded using a low-end webcam. \textit{Caltech} is similar to \textit{Amazon} but with various viewpoint variations. The 10 categories include \texttt{backpack}, \texttt{bike}, \texttt{calculator}, \texttt{headphones}, \texttt{keyboard}, \texttt{laptop} \texttt{computer}, \texttt{monitor}, \texttt{mouse}, \texttt{mug}, and \texttt{projector}. Some images of four domains are shown in Fig.~\ref{fig:offcal10}. Overall, we have about 2500 images and 12 domain adaptation problems. For each problem, we repeat the experiment 20 times. In each trail, we randomly select 20 images from each category for training if the domain is \textit{Amazon} and \textit{Caltech}, or 8 images if the domain is \textit{DSLR} or \textit{web-cam}. All images in the target domain are employed in the both adaptation and testing stages. The mean and standard deviation of multi-class accuracy are reported.

\begin{figure}[!t]
	\centering
	\includegraphics[width=14cm]{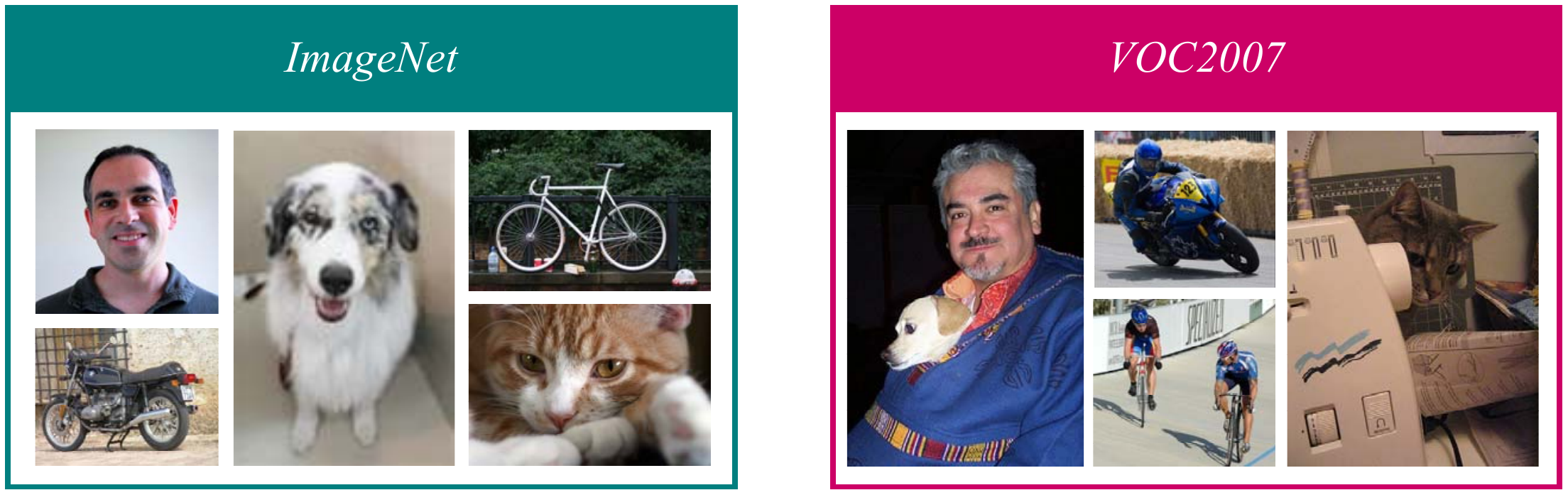}
	\caption{Some images from ImageNet-VOC2007 dataset. 5 categories of \texttt{person}, \texttt{dog}, \texttt{motorbike}, \texttt{bicycle}, and \texttt{cat} are presented.}
	\label{fig:imagenet-voc2007}
\end{figure}

\paragraph{Office dataset.} Office dataset is developed by~\cite{saenko2010adapting} and turns out to be a standard benchmark for the evaluation of domain adaptation. It consists of 31 categories and 3 domains, leading to 6 domain adaptation problems. Among these 31 categories, only 16 overlap with the categories contained in the 1000-category ImageNet 2012 dataset\footnote{The 16 overlapping categories are \texttt{backpack}, \texttt{bike helmet}, \texttt{bottle}, \texttt{desk lamp}, \texttt{desk computer}, \texttt{file cabinet}, \texttt{keyboard}, \texttt{laptop computer}, \texttt{mobile phone}, \texttt{mouse}, \texttt{printer}, \texttt{projector}, \texttt{ring binder}, \texttt{ruler}, \texttt{speaker}, and \texttt{trash can}.}~\cite{hoffman2013one}, so Office dataset is more challenging than its counterpart Office-Caltech10 dataset.  We follow the same experimental protocol mentioned above to conduct the experiments, so in each task we have 620 images in all from the source domain.

\paragraph{ImageNet--VOC2007 dataset.} As described in the main text, ImageNet and VOC 2007 datasets are used to evaluate the domain adaptation performance from single-label to multi-label situation. The same 20 categories as the VOC 2007 dataset are chosen from original ImageNet dataset. These 20 categories are \texttt{aeroplane}, \texttt{bicycle}, \texttt{bird}, \texttt{boat}, \texttt{bottle}, \texttt{bus}, \texttt{car}, \texttt{cat}, \texttt{chair}, \texttt{cow}, \texttt{dining table}, \texttt{dog}, \texttt{horse}, \texttt{motorbike}, \texttt{person}, \texttt{potted plant}, \texttt{sheep}, \texttt{sofa}, \texttt{train}, and \texttt{tv monitor}. The 20-category ImageNet subset is adopted as the source domain, and the \texttt{test} subset of VOC2007 is employed as the target domain. Some images of two domains are illustrated in Fig.~\ref{fig:painting}. Also, the similar experimental protocol mentioned above is used. The difference, however, is that we report the mean and standard deviation of average precision (AP) for each category, respectively.

\section{Recognition results}
\label{sec:results}

We compare against the same methods used in the main text, including the baseline no adaptation (NA), principal components analysis (PCA), transfer component analysis (TCA)~\cite{Sinno11TCA}, geodesic flow kernel (GFK)~\cite{gong12geodesic}, domain-invariant projection (DIP)~\cite{baktashmotlagh2013dip}, subspace alignment (SA)~\cite{fernando13sa}, low-rank transfer subspace learning (LTSL)~\cite{shao2014ltsl}, landmarks selection subspace alignment (LSSA)~\cite{aljundi2015landmarks}, and correlation alignment (CORAL)~\cite{sun2016return}. Our approach is denoted by NTSL (the naive version) and TAISL. We also extract convolutional activations from the CONV5\_3 layer of the VGG--VD--16 model~\cite{Simonyan14verydeep}. We mark the feature as \textsc{vCONV} and \textsc{tCONV} for vectorized and tensor-form convolutional activations, respectively. The same parameters described in the main text are set to report the results.

\paragraph{Office results.} Results of the Office dataset are listed in Table~\ref{tab:off31_vgg_vd_16_res}. Similar to the tendency shown by the results of Office-Caltech10 dataset in the main text, our approach outperforms or is on par with other comparing methods. It is interesting that sometimes NTSL even achieves better results than TAISL. We believe such results are sound, because a blind global adaptation cannot always achieve accuracy improvement. However, it is clear that learning an invariant tensor space works much better than learning a shared vector space. Furthermore, the joint learning effectively reduces the standard deviation and thus improves the stability of the adaptation.

\paragraph{ImageNet--VOC2007 results.} Table~\ref{tab:imagenet_voc_vgg_vd_16_res} shows the complete results on ImageNet--VOC2007 dataset (only partial results are presented in the main text due to the page limitation). Our approach achieves the best mean accuracy in 4 and the second best in 6 out of 20 categories. In general, when noisy labels exist in the target domain, our approach demonstrates a stable improvement in accuracy. Moreover, compared to the baseline NTSL, the standard deviation is generally reduced, which means aligning the source domain to the target not only promotes the classification accuracy but also improves the stability of tensor space.

\begin{table*}[!tb]
	\centering
	\addtolength{\tabcolsep}{4.5pt}
	\renewcommand\arraystretch{1.25}
	\begin{tabular}{l|c|cccccccc}
		\hline
		Method & Feature & \small A$\rightarrow$D  & \small D$\rightarrow$A  & \small A$\rightarrow$W  & \small W$\rightarrow$A  & \small D$\rightarrow$W  & \small W$\rightarrow$D  & \textsc{mean}\\
		\hline
		NA    & \textsc{vCONV} & 53.8\small({2.3}) & 39.3\small({1.7}) & 47.7\small({1.7}) & 36.3\small({1.6}) & 77.4\small({1.7}) & 81.3\small({1.5}) & 56.0\\
		PCA   & \textsc{vCONV} & 40.5\small({3.3}) & 38.2\small({2.6}) & 36.5\small({2.9}) & 37.8\small({2.9}) & 68.7\small({2.5}) & 70.5\small({2.6}) & 48.7\\
		DAUME & \textsc{vCONV} & 48.4\small({2.5}) & 35.2\small({1.5}) & 42.5\small({2.0}) & 33.6\small({1.8}) & 68.4\small({2.5}) & 74.2\small({2.4}) & 50.4\\
		TCA   & \textsc{vCONV} & 30.3\small({4.5}) & 20.1\small({4.4}) & 27.0\small({3.1}) & 18.1\small({3.0}) & 51.1\small({3.2}) & 53.0\small({3.2}) & 33.3\\
		GFK   & \textsc{vCONV} & 47.4\small({4.7}) & 36.2\small({2.9}) & 41.5\small({3.5}) & 33.4\small({2.6}) & 75.3\small({1.6}) & 78.0\small({2.4}) & 51.9\\
		DIP   & \textsc{vCONV} & 36.8\small({4.5}) & 13.8\small({1.8}) & 29.6\small({5.0}) & 17.8\small({2.6}) & 77.4\small({1.8}) & 81.5\small({2.0}) & 42.8\\
		SA    & \textsc{vCONV} & 28.6\small({3.5}) & 37.1\small({2.1}) & 29.0\small({2.1}) & 34.9\small({2.9}) & 75.1\small({2.4}) & 75.1\small({2.7}) & 46.6\\
		LTSL  & \textsc{vCONV} & 32.0\small({5.5}) & 28.6\small({1.6}) & 24.2\small({3.7}) & 27.1\small({2.0}) & 60.9\small({4.0}) & 73.9\small({3.3}) & 41.1\\
		LSSA  & \textsc{vCONV} & \textbf{56.6\small({2.0})} & 45.6\small({1.6}) & \textbf{52.2\small({1.6})} & 40.7\small({2.0}) & 73.0\small({2.1}) & 63.5\small({3.8}) & 55.3\\
		CORAL & \textsc{vCONV} & 39.9\small({1.7}) & 42.7\small({0.9}) & 39.7\small({1.7}) & 40.7\small({1.0}) & 82.0\small({1.3}) & 79.5\small({1.4}) & 54.1\\
		\hline
		NTSL   & \textsc{tCONV} & 56.1\small({2.4}) & \red{45.7\small({1.5})} & \red{50.8\small({2.3})} & \red{42.6\small({2.2})} & \red{84.4\small({1.6})} & \red{88.2\small({1.4})} & \red{61.3}\\
		TAISL  & \textsc{tCONV} & \red{56.4\small({2.4})} & \textbf{45.9\small({1.1})} & 50.7\small({2.0}) & \textbf{43.2\small({1.7})} & \textbf{84.5\small({1.4})} & \textbf{88.5\small({1.2})} & \textbf{61.5}\\
		\hline
	\end{tabular}
	\caption{Average multi-class recognition accuracy (\%) on Office dataset over 20 trials. The highest accuracy in each column is boldfaced, the second best is marked in red, and standard deviations are shown in parentheses.}
	\label{tab:off31_vgg_vd_16_res}
\end{table*}

\begin{table*}[!tb] \footnotesize
	\centering
	\addtolength{\tabcolsep}{-1pt}
	\renewcommand\arraystretch{1.35}
	\begin{tabular}{l|c|cccccccccccc}
		\hline
		\multicolumn{2}{c|}{VOC 2007 test} & \footnotesize aero  & \footnotesize bike  & \footnotesize bird  & \footnotesize boat  & \footnotesize bottle  & \footnotesize bus  & \footnotesize car  & \footnotesize cat  & \footnotesize chair  & \footnotesize cow\\
		\hline
		NA    & \textsc{vCONV} & 66.4\sz({2.1})  & 65.3\sz({2.9})  & 65.6\sz({4.0})  & 56.1\sz({9.0})  & 29.5\sz({2.1})  & 51.2\sz({3.4})  & 70.9\sz({4.5})  & 70.6\sz({3.4})  & 19.3\sz({2.0})  & 30.3\sz({8.0}) \\
		PCA   & \textsc{vCONV} & 28.9\sz({5.8})  & 25.3\sz({7.2})  & 30.2\sz({3.9})  & 14.0\sz({4.8})  & 23.3\sz({5.2})  & 15.6\sz({6.3})  & 41.5\sz({7.5})  & 44.9\sz({4.5})  & 11.2\sz({0.9})  & 6.0\sz({1.8}) \\
		Daum\'{e} III & \textsc{vCONV} & 64.1\sz({3.7})  & 60.4\sz({4.2})  & 59.7\sz({7.4})  & 53.5\sz({7.8})  & 26.6\sz({3.3})  & 49.0\sz({5.1})  & 66.3\sz({5.1})  & 65.7\sz({5.3})  & 18.6\sz({3.5})  & 26.9\sz({8.5}) \\
		TCA   & \textsc{vCONV} & 43.2\sz({9.8})  & 46.0\sz({17.0})  & 44.4\sz({10.5})  & 25.3\sz({13.0})  & 20.7\sz({1.7})  & 30.4\sz({7.7})  & 59.5\sz({8.6})  & 56.7\sz({8.2})  & 17.1\sz({3.0})  & 16.9\sz({6.3}) \\
		GFK   & \textsc{vCONV} & 70.0\sz({6.9})  & \textbf{66.0\sz({7.6})}  & 74.6\sz({3.8})  & 40.7\sz({11.8})  & 32.5\sz({4.4})  & 55.0\sz({6.9})  & \red{71.3\sz({5.2})}  & 73.1\sz({6.5})  & 16.3\sz({3.6})  & 28.9\sz({5.3}) \\
		DIP   & \textsc{vCONV} & 69.8\sz({5.5})  & \red{65.8\sz({7.2})}  & \red{78.4\sz({4.6})}  & 34.2\sz({9.1})  & 29.1\sz({5.0})  & 54.4\sz({7.3})  & \textbf{75.7\sz({3.9})}  & \red{75.9\sz({3.7})}  & \red{20.1\sz({4.6})}  & 25.5\sz({5.0}) \\
		SA    & \textsc{vCONV} & 64.4\sz({10.1})  & 54.4\sz({9.3})  & 69.3\sz({5.4})  & 50.8\sz({12.7})  & 34.4\sz({4.6})  & 50.8\sz({6.5})  & 64.3\sz({9.5})  & 67.4\sz({4.9})  & 11.2\sz({1.9})  & 18.4\sz({6.6}) \\
		LTSL  & \textsc{vCONV} & 56.9\sz({10.4})  & 59.8\sz({6.3})  & 61.0\sz({7.7})  & 50.6\sz({15.6})  & 34.9\sz({6.2})  & 50.9\sz({9.6})  & 66.9\sz({3.6})  & 70.8\sz({8.8})  & 11.4\sz({1.5})  & 21.9\sz({6.3}) \\
		LSSA & \textsc{vCONV} & \textbf{78.7\sz({2.0})} & 71.8\sz({1.5}) & \textbf{79.7\sz({1.2})} & 18.5\sz({2.0}) & \textbf{38.4\sz({4.6})} & \textbf{64.1\sz({3.2})} & 69.4\sz({2.2}) & \textbf{81.7\sz({0.5})} & \textbf{57.2\sz({2.4})} & 29.5\sz({1.9}) \\
		CORAL & \textsc{vCONV} & 71.4\sz({3.3})  & 63.3\sz({4.3})  & 71.7\sz({3.6})  & 58.6\sz({9.5})  & 35.2\sz({2.4})  & 61.9\sz({3.6})  & 62.7\sz({7.1})  & 72.0\sz({4.3})  & 18.7\sz({2.7})  & \textbf{36.0\sz({5.7})}\\
		\hline
		NTSL   & \textsc{tCONV} & 76.3\sz({4.3})  & 61.6\sz({5.5})  & 71.0\sz({3.9})  & \textbf{65.9\sz({8.3})}  & 35.7\sz({3.7})  & 56.1\sz({7.1})  & 70.1\sz({4.8})  & 71.3\sz({3.2})  & 16.6\sz({2.6})  & \red{34.7\sz({9.8})}\\
		TAISL  & \textsc{tCONV} & \red{76.4\sz({5.1})}  & 62.3\sz({4.8})  & 71.6\sz({3.1})  & \red{64.9\sz({7.7})}  & \red{36.7\sz({3.5})}  & 57.0\sz({6.6})  & 71.2\sz({4.3})  & 72.0\sz({2.1})  & 15.7\sz({2.9})  & 33.3\sz({6.6})\\
		\hline
		\multicolumn{2}{c|}{} & \footnotesize table  & \footnotesize dog  & \footnotesize horse  & \footnotesize mbike  & \footnotesize person  & \footnotesize plant  & \footnotesize sheep  & \footnotesize sofa  & \footnotesize train  & \footnotesize tv & mAP\\
		\hline
		NA    & \textsc{vCONV} & 35.7\sz({5.5}) & 47.9\sz({6.4}) & 35.5\sz({11.4}) & 47.0\sz({8.0}) & 69.3\sz({2.9}) & \red{25.6\sz({3.9})} & 44.9\sz({6.9}) & 46.9\sz({5.3}) & 71.8\sz({4.4}) & 56.4\sz({3.3}) & 50.3\\
		PCA   & \textsc{vCONV} & 29.0\sz({6.9}) & 32.5\sz({4.6}) & 23.2\sz({6.2}) & 25.0\sz({5.0}) & 70.2\sz({1.9}) & 9.3\sz({4.3}) & 11.7\sz({3.5}) & 16.2\sz({2.8}) & 29.0\sz({6.7}) & 29.0\sz({6.6}) & 25.8\\
		Daum\'{e} III & \textsc{vCONV} & 30.0\sz({5.4}) & 43.6\sz({6.9}) & 28.3\sz({8.7}) & 40.5\sz({6.6}) & 68.5\sz({2.5}) & 23.6\sz({3.5}) & 37.7\sz({7.4}) & 44.5\sz({5.6}) & 67.6\sz({5.4}) & 51.9\sz({4.4}) & 46.4\\
		TCA   & \textsc{vCONV} & 27.6\sz({8.7}) & 43.2\sz({7.6}) & 29.0\sz({14.6}) & 31.8\sz({10.2}) & 58.1\sz({5.7}) & 11.6\sz({4.5}) & 22.7\sz({8.0}) & 24.0\sz({9.4}) & 52.3\sz({8.9}) & 33.6\sz({10.2}) & 34.7\\
		GFK   & \textsc{vCONV} & 48.3\sz({10.4}) & 56.7\sz({7.4}) & \red{59.2\sz({16.7})} & 58.3\sz({4.8}) & \textbf{75.8\sz({3.6})} & 15.2\sz({4.6}) & 52.5\sz({4.8}) & 44.7\sz({6.0}) & \red{79.9\sz({4.9})} & 57.1\sz({4.5}) & 53.8\\
		DIP   & \textsc{vCONV} & 42.2\sz({8.1}) & 53.7\sz({5.4}) & \textbf{64.7\sz({7.1})} & 56.3\sz({5.7}) & \red{73.5\sz({3.1})} & 14.7\sz({4.2}) & 48.9\sz({4.3}) & 39.8\sz({10.0}) & \textbf{80.5\sz({5.6})} & 59.4\sz({5.2}) & 53.1\\
		SA    & \textsc{vCONV} & 36.9\sz({12.8}) & 54.2\sz({5.7}) & 39.5\sz({15.5}) & 53.7\sz({10.9}) & 68.9\sz({2.4}) & 20.9\sz({6.7}) & 31.4\sz({10.2}) & 29.3\sz({6.1}) & 73.5\sz({5.2}) & 55.2\sz({5.7}) & 47.4\\
		LTSL  & \textsc{vCONV} & 43.7\sz({12.4}) & 55.4\sz({7.4}) & 53.4\sz({13.1}) & 52.5\sz({10.7}) & 69.9\sz({4.3}) & 18.8\sz({8.2}) & 38.2\sz({9.5}) & 28.9\sz({13.2}) & 67.1\sz({9.9}) & 54.0\sz({7.5}) & 48.3\\
		LSSA & \textsc{vCONV} & 33.7\sz({3.4}) & 56.9\sz({2.5}) & 41.1\sz({5.4}) & 56.3\sz({9.3}) & 51.2\sz({2.0}) & 15.3\sz({5.7}) & 32.5\sz({10.6}) & 43.4\sz({8.4}) & 81.1\sz({1.4}) & 51.6\sz({4.6}) & 52.6\\
		CORAL & \textsc{vCONV} & 40.6\sz({6.7}) & 53.8\sz({5.3}) & 34.8\sz({6.8}) & 57.3\sz({5.6}) & 67.6\sz({2.0}) & 24.2\sz({1.5}) & \textbf{54.8\sz({2.9})} & 47.7\sz({6.2}) & 71.7\sz({3.5}) & 56.9\sz({3.6}) & 53.0\\
		\hline
		NTSL   & \textsc{tCONV} & \red{49.8\sz({10.4})} & \textbf{58.3\sz({5.1})} & 40.9\sz({12.9}) & \red{59.7\sz({10.2})} & 72.0\sz({4.6}) & 25.0\sz({4.5}) & 53.4\sz({6.0}) & \textbf{49.8\sz({4.6})} & 75.3\sz({3.6}) & \red{60.2\sz({3.5})} & \red{55.2}\\
		TAISL  & \textsc{tCONV} & \textbf{50.7\sz({10.0})} & \red{57.6\sz({3.8})} & 39.0\sz({14.0}) & \textbf{60.3\sz({8.7})}& 72.2\sz({3.8}) & \textbf{26.6\sz({5.4})} & \red{53.6\sz({5.6})} & \red{49.8\sz({5.6})} & 74.2\sz({4.9}) & \textbf{60.4\sz({3.5})} & \textbf{55.3}\\
		\hline
	\end{tabular}
	\caption{Average precision (\%) on ImageNet-VOC2007 dataset over 10 trials. The highest AP in each column is boldfaced, the second best is marked in red, and standard deviations are shown in parentheses.}
	\label{tab:imagenet_voc_vgg_vd_16_res}
\end{table*}

\begin{figure*}[!tb]
	\centering
	\subfloat{\includegraphics[width=0.33\linewidth]{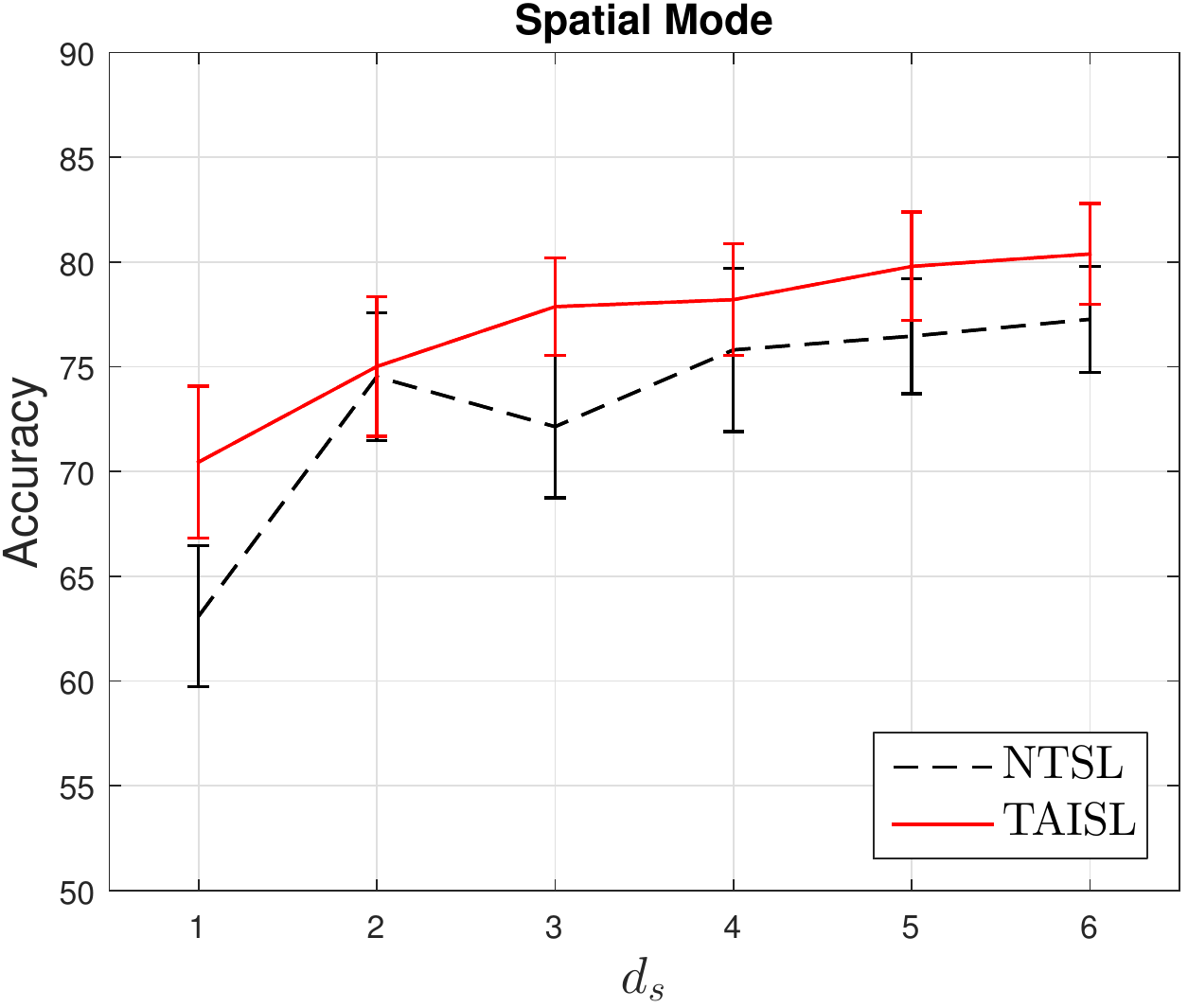}}
	\subfloat{\includegraphics[width=0.33\linewidth]{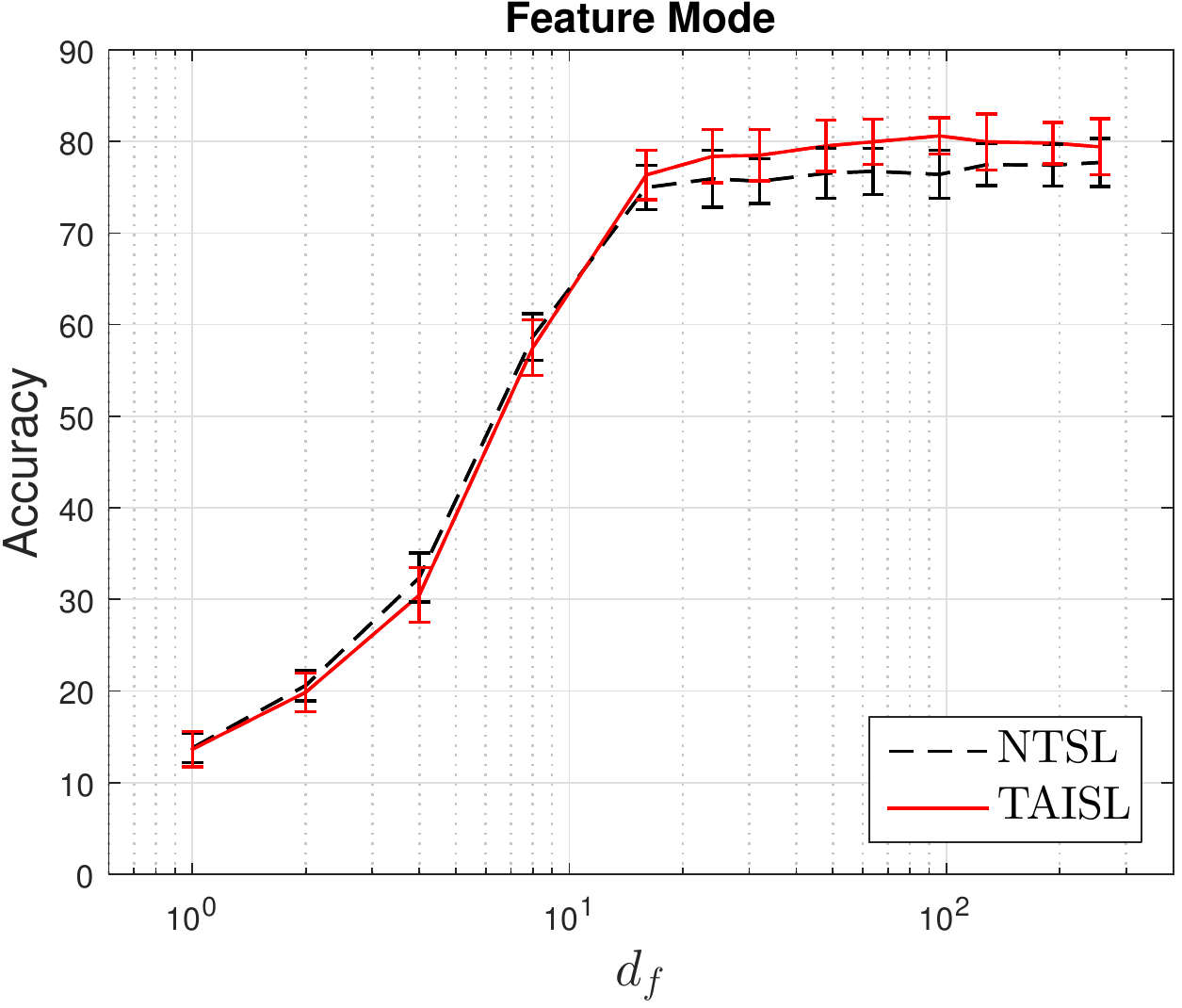}}
	\subfloat{\includegraphics[width=0.335\linewidth]{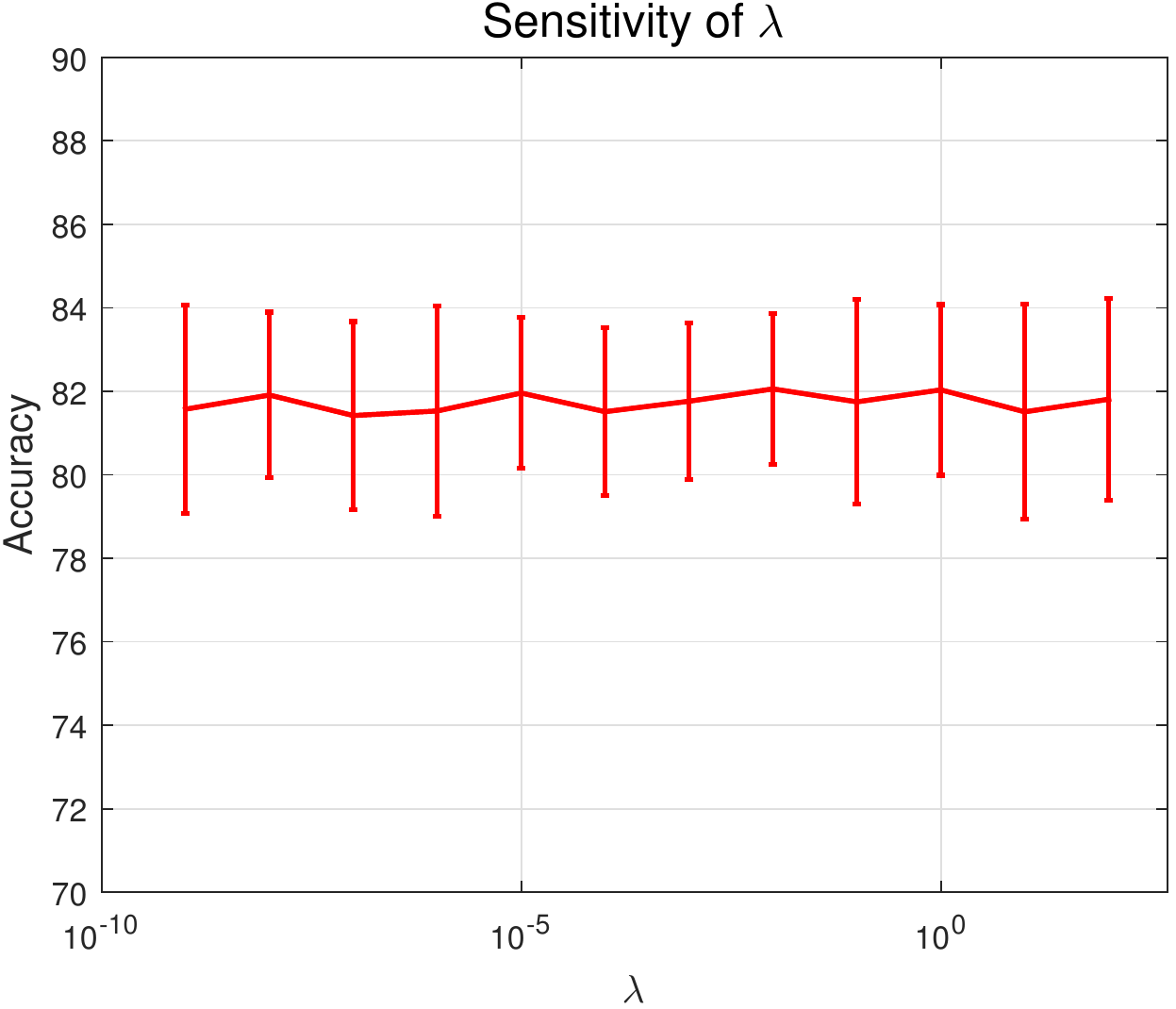}}
	\captionsetup{singlelinecheck=false}
	\caption{Sensitivity of tensor subspace dimensionality and weight coefficient $\lambda$ on the DA task of W$\rightarrow$C.}
	\label{fig:param}
\end{figure*}

\section{Parameters Sensitivity}
\label{sec:sensitivity}

Here we investigate the sensitivity of 3 parameters involved in our approach. Specifically, they are the spatial mode dimensionality $d_s$ ($d_1$ and $d_2$ in the main text, we assume $d_1=d_2=d_s$), the feature mode dimensionality $d_f$ ($d_3$ in the main text), and the weight coefficient $\lambda$. We monitor how the classification accuracy changes when these parameters vary. At each time, only one parameter is allowed to change. By default, $d_s=6$, $d_f=128$, and $\lambda=1e^{-5}$. A DA task of W$\rightarrow$C from the Office-Caltech10 dataset is chosen. Results are illustrated by Fig.~\ref{fig:param}. According to Fig.~\ref{fig:param}, we can make the following observations:
\squishlist
\item In general, there exhibits a tendency for increased $d_s$ to increased classification accuracy, which implies that the adaptation can benefit from extra spatial information. This is why we preserve the original spatial mode as it is.
\item As per the feature mode dimensionality $d_f$, a dramatic growth appears when $d_f$ increases from $1$ to $16$. However, the classification accuracy starts to level off when $d_f$ exceeds $16$. Such results make sense, because when the feature dimensionality is relatively small, the discriminative power of feature representations cannot be guaranteed. Overall, our approach demonstrates stable classification performance over a wide range of feature mode dimensionality.
\item Only a slight fluctuation occurs when $\lambda$ varies between $1e^{-9}$ and $1e^1$. The classification accuracy is virtually insensitive to the weight coefficient $\lambda$. This is another good property of our approach.
\squishend


\paragraph{Acknowledgments.} This work was supported in part by the National High-tech R\&D Program of China (863 Program) under Grant 2015AA015904 and in part by the National Natural Science Foundation of China under Grant 61502187.

{\small
	\bibliographystyle{ieee}
	\bibliography{refbib}

\begin{thebibliography}{10}\itemsep=-1pt

\bibitem{aljundi2015landmarks}
R.~Aljundi, R.~Emonet, D.~Muselet, and M.~Sebban.
\newblock Landmarks-based kernelized subspace alignment for unsupervised domain
  adaptation.
\newblock In {\em Proc. IEEE Conference on Computer Vision and Pattern
  Recognition (CVPR)}, pages 56--63, 2015.

\bibitem{baktashmotlagh2013dip}
M.~Baktashmotlagh, M.~T. Harandi, B.~C. Lovell, and M.~Salzmann.
\newblock Unsupervised domain adaptation by domain invariant projection.
\newblock In {\em Proc. IEEE International Conference on Computer Vision
  (ICCV)}, pages 769--776, 2013.

\bibitem{ben2007analysis}
S.~Ben-David, J.~Blitzer, K.~Crammer, F.~Pereira, et~al.
\newblock Analysis of representations for domain adaptation.
\newblock In {\em Advances in Neural Information Processing Systems (NIPS)},
  volume~19, page 137, 2007.

\bibitem{Chatfield14}
K.~Chatfield, K.~Simonyan, A.~Vedaldi, and A.~Zisserman.
\newblock Return of the devil in the details: Delving deep into convolutional
  nets.
\newblock In {\em Proc. British Machine Vision Conference (BMVC)}, 2014.

\bibitem{Chu2013STM}
W.-S. Chu, F.~De~La~Torre, and J.~F. Cohn.
\newblock Selective transfer machine for personalized facial action unit
  detection.
\newblock In {\em Proc. IEEE Conference on Computer Vision and Pattern
  Recognition (CVPR)}, June 2013.

\bibitem{daume2007frustratingly}
H.~Daum{\'e}~III.
\newblock Frustratingly easy domain adaptation.
\newblock In {\em Proc. Association for Computational Linguistics (ACL)}, 2007.

\bibitem{Deng2009}
J.~Deng, W.~Dong, R.~Socher, L.-J. Li, K.~Li, and L.~Fei-Fei.
\newblock Imagenet: A large-scale hierarchical image database.
\newblock In {\em Proc. IEEE Conference on Computer Vision and Pattern
  Recognition (CVPR)}, pages 248--255, 2009.

\bibitem{everingham2010pascal}
M.~Everingham, L.~Van~Gool, C.~K. Williams, J.~Winn, and A.~Zisserman.
\newblock The pascal visual object classes ({VOC}) challenge.
\newblock {\em International Journal of Computer Vision}, 88(2):303--338, 2010.

\bibitem{felzenszwalb2010dpm}
P.~F. Felzenszwalb, R.~B. Girshick, D.~McAllester, and D.~Ramanan.
\newblock Object detection with discriminatively trained part-based models.
\newblock {\em IEEE Transactions on Pattern Analysis and Machine Intelligence},
  32(9):1627--1645, 2010.

\bibitem{fernando13sa}
B.~Fernando, A.~Habrard, M.~Sebban, and T.~Tuytelaars.
\newblock Unsupervised visual domain adaptation using subspace alignment.
\newblock In {\em Proc. IEEE International Conference on Computer Vision
  (ICCV)}, pages 2960--2967, 2013.

\bibitem{ganin2015unsupervised}
Y.~Ganin and V.~Lempitsky.
\newblock Unsupervised domain adaptation by backpropagation.
\newblock In {\em Proc. International Conference on Machine Learning (ICML)},
  pages 1180--1189, 2015.

\bibitem{gong12geodesic}
B.~Gong, Y.~Shi, F.~Sha, and K.~Grauman.
\newblock Geodesic flow kernel for unsupervised domain adaptation.
\newblock In {\em Proc. IEEE Conference on Computer Vision and Pattern
  Recognition (CVPR)}, pages 2066--2073, 2012.

\bibitem{gopalan2011domain}
R.~Gopalan, R.~Li, and R.~Chellappa.
\newblock Domain adaptation for object recognition: An unsupervised approach.
\newblock In {\em Proc. IEEE International Conference on Computer Vision
  (ICCV)}, pages 999--1006, 2011.

\bibitem{He2015}
K.~He, X.~Zhang, S.~Ren, and J.~Sun.
\newblock Spatial pyramid pooling in deep convolutional networks for visual
  recognition.
\newblock {\em IEEE Transactions on Pattern Analysis and Machine Intelligence},
  37(9):1904--1916, 2015.

\bibitem{he2016deep}
K.~He, X.~Zhang, S.~Ren, and J.~Sun.
\newblock Deep residual learning for image recognition.
\newblock In {\em Proc. IEEE Conference on Computer Vision and Pattern
  Recognition (CVPR)}, 2016.

\bibitem{hoffman2013one}
J.~Hoffman, E.~Tzeng, J.~Donahue, Y.~Jia, K.~Saenko, and T.~Darrell.
\newblock One-shot adaptation of supervised deep convolutional models.
\newblock In {\em Proc. International Conference on Learning Representations
  Workshops (ICLRW)}, 2013.

\bibitem{Kim09TCCA}
T.~K. Kim and R.~Cipolla.
\newblock Canonical correlation analysis of video volume tensors for action
  categorization and detection.
\newblock {\em IEEE Transactions on Pattern Analysis and Machine Intelligence},
  31(8):1415--1428, Aug 2009.

\bibitem{kolda2009tensor}
T.~G. Kolda and B.~W. Bader.
\newblock Tensor decompositions and applications.
\newblock {\em SIAM review}, 51(3):455--500, 2009.

\bibitem{Krizhevsky2012}
A.~Krizhevsky, I.~Sutskever, and G.~E. Hinton.
\newblock Imagenet classification with deep convolutional neural networks.
\newblock In {\em Advances in Neural Information Processing Systems (NIPS)},
  pages 1097--1105, 2012.

\bibitem{li2007robust}
X.~Li, W.~Hu, Z.~Zhang, X.~Zhang, and G.~Luo.
\newblock Robust visual tracking based on incremental tensor subspace learning.
\newblock In {\em Proc. IEEE International Conference on Computer Vision
  (ICCV)}, pages 1--8. IEEE, 2007.

\bibitem{long15fcn}
J.~Long, E.~Shelhamer, and T.~Darrell.
\newblock Fully convolutional networks for semantic segmentation.
\newblock In {\em 2015 IEEE Conference on Computer Vision and Pattern
  Recognition (CVPR)}, pages 3431--3440, June 2015.

\bibitem{lu20172dsa}
H.~Lu, Z.~Cao, Y.~Xiao, and Y.~Zhu.
\newblock Two-dimensional subspace alignment for convolutional activations
  adaptation.
\newblock {\em Pattern Recognition}, 71:320--336, 2017.

\bibitem{Sinno11TCA}
S.~J. Pan, I.~W. Tsang, J.~T. Kwok, and Q.~Yang.
\newblock Domain adaptation via transfer component analysis.
\newblock {\em IEEE Transactions on Neural Networks}, 22(2):199--210, Feb 2011.

\bibitem{patel2015visual}
V.~M. Patel, R.~Gopalan, R.~Li, and R.~Chellappa.
\newblock Visual domain adaptation: A survey of recent advances.
\newblock {\em IEEE Signal Processing Magazine}, 32(3):53--69, 2015.

\bibitem{ren2015faster}
S.~Ren, K.~He, R.~Girshick, and J.~Sun.
\newblock Faster {R-CNN}: Towards real-time object detection with region
  proposal networks.
\newblock In {\em Proc. Advances in Neural Information Processing Systems
  (NIPS)}, pages 91--99, 2015.

\bibitem{saenko2010adapting}
K.~Saenko, B.~Kulis, M.~Fritz, and T.~Darrell.
\newblock Adapting visual category models to new domains.
\newblock In {\em Proc. European Conference on Computer Vision (ECCV)}, pages
  213--226, 2010.

\bibitem{sener2016learning}
O.~Sener, H.~O. Song, A.~Saxena, and S.~Savarese.
\newblock Learning transferrable representations for unsupervised domain
  adaptation.
\newblock In {\em Advances in Neural Information Processing Systems (NIPS)},
  pages 2110--2118, 2016.

\bibitem{shakhnarovich2011face}
G.~Shakhnarovich and B.~Moghaddam.
\newblock Face recognition in subspaces.
\newblock In {\em Handbook of Face Recognition}, pages 19--49. Springer, 2011.

\bibitem{shao2014ltsl}
M.~Shao, D.~Kit, and Y.~Fu.
\newblock Generalized transfer subspace learning through low-rank constraint.
\newblock {\em International Journal of Computer Vision}, 109(1-2):74--93,
  2014.

\bibitem{Simonyan14verydeep}
K.~Simonyan and A.~Zisserman.
\newblock Very deep convolutional networks for large-scale image recognition.
\newblock {\em CoRR}, abs/1409.1556, 2014.

\bibitem{sun2016return}
B.~Sun, J.~Feng, and K.~Saenko.
\newblock Return of frustratingly easy domain adaptation.
\newblock In {\em Proc. AAAI Conference on Artificial Intelligence}, 2016.

\bibitem{van2008visualizing}
L.~Van~der Maaten and G.~Hinton.
\newblock Visualizing data using {t-SNE}.
\newblock {\em Journal of Machine Learning Research}, 9(2579--2605):85, 2008.

\bibitem{vasilescu2002multilinear}
M.~A.~O. Vasilescu and D.~Terzopoulos.
\newblock Multilinear analysis of image ensembles: Tensorfaces.
\newblock In {\em Proc. European Conference on Computer Vision (ECCV)}, pages
  447--460. Springer, 2002.

\bibitem{vasilescu2003multilinear}
M.~A.~O. Vasilescu and D.~Terzopoulos.
\newblock Multilinear subspace analysis of image ensembles.
\newblock In {\em Proc. IEEE Conference on Computer Vision and Pattern
  Recognition (CVPR)}, volume~2, pages II--93. IEEE, 2003.

\bibitem{wang2015beyond}
L.~Wang, J.~Zhang, L.~Zhou, C.~Tang, and W.~Li.
\newblock Beyond covariance: Feature representation with nonlinear kernel
  matrices.
\newblock In {\em Proc. IEEE International Conference on Computer Vision
  (ICCV)}, pages 4570--4578, 2015.

\bibitem{wen2016discriminative}
Y.~Wen, K.~Zhang, Z.~Li, and Y.~Qiao.
\newblock A discriminative feature learning approach for deep face recognition.
\newblock In {\em Proc. European Conference on Computer Vision (ECCV)}, pages
  499--515. Springer, 2016.

\bibitem{wen2013feasible}
Z.~Wen and W.~Yin.
\newblock A feasible method for optimization with orthogonality constraints.
\newblock {\em Mathematical Programming}, 142(1-2):397--434, 2013.

\bibitem{yosinski2014transferable}
J.~Yosinski, J.~Clune, Y.~Bengio, and H.~Lipson.
\newblock How transferable are features in deep neural networks?
\newblock In {\em Advances in Neural Information Processing Systems (NIPS)},
  pages 3320--3328, 2014.

\bibitem{zhang2016exploring}
L.~Zhang, W.~Wei, C.~Tian, F.~Li, and Y.~Zhang.
\newblock Exploring structured sparsity by a reweighted laplace prior for
  hyperspectral compressive sensing.
\newblock {\em IEEE Transactions on Image Processing}, 25(10):4974--4988, 2016.

\bibitem{zhang2016dictionary}
L.~Zhang, W.~Wei, Y.~Zhang, C.~Shen, A.~van~den Hengel, and Q.~Shi.
\newblock Dictionary learning for promoting structured sparsity in
  hyperspectral compressive sensing.
\newblock {\em IEEE Transactions on Geoscience and Remote Sensing},
  54(12):7223--7235, 2016.

\bibitem{Zhao15CP}
Q.~Zhao, L.~Zhang, and A.~Cichocki.
\newblock Bayesian {CP} factorization of incomplete tensors with automatic rank
  determination.
\newblock {\em IEEE Transactions on Pattern Analysis and Machine Intelligence},
  37(9):1751--1763, Sept 2015.

\end{thebibliography}
}

\end{document}